\documentclass[letterpaper]{article} 
\usepackage{aaai2026}  
\usepackage{times}  
\usepackage{helvet}  
\usepackage{courier}  
\usepackage[hyphens]{url}  
\usepackage{graphicx} 
\usepackage{natbib}  
\usepackage{caption} 
\usepackage{algorithm}
\usepackage{algorithmic}
\usepackage{multirow}
\usepackage{graphicx}
\usepackage{newfloat}
\usepackage{listings}
\usepackage{booktabs}
\usepackage[utf8]{inputenc}
\usepackage{times}
\usepackage{amsfonts} 
\usepackage{bm} 
\usepackage{color}
\usepackage{graphicx}
\usepackage{amsmath, amssymb} 
\DeclareCaptionStyle{ruled}{labelfont=normalfont,labelsep=colon,strut=off} 
\floatstyle{ruled}
\newfloat{listing}{tb}{lst}{}
\floatname{listing}{Listing}
\nocopyright
\title{EvolvR: Self-Evolving Pairwise Reasoning for Story Evaluation to Enhance Generation}

\author {
    Xinda Wang\textsuperscript{\rm \dag, 1},
    Zhengxu Hou\textsuperscript{\rm \dag, 2},
    Yangshijie Zhang\textsuperscript{\rm 3}, 
    Bingren Yan\textsuperscript{\rm 2}, 
    Jialin Liu\textsuperscript{\rm 1}, \\
    Chenzhuo Zhao\textsuperscript{\rm 1},
    Zhibo Yang\textsuperscript{\rm 2},
    Bin-Bin Yang\textsuperscript{\rm 2},
    Feng Xiao\textsuperscript{\rm *, 2}
}
\affiliations {
    \textsuperscript{\rm 1}Peking University \qquad 
    \textsuperscript{\rm 2}Alibaba Group \\
    \textsuperscript{\rm 3}Lanzhou University \qquad  
    \textsuperscript{\dag}Contributed equally. \qquad 
    \textsuperscript{*}Corresponding author. \\ 
}

\usepackage{bibentry}

\begin{document}
\maketitle

\begin{abstract}
Although the effectiveness of Large Language Models (LLMs) as judges (LLM-as-a-judge) has been validated, their performance remains limited in open-ended tasks, particularly in story evaluation. Accurate story evaluation is crucial not only for assisting human quality judgment but also for providing key signals to guide story generation. However, existing methods face a dilemma: prompt engineering for closed-source models suffers from poor adaptability, while fine-tuning approaches for open-source models lack the rigorous reasoning capabilities essential for story evaluation. To address this, we propose the Self-Evolving Pairwise Reasoning (EvolvR) framework. Grounded in pairwise comparison, the framework first self-synthesizes score-aligned Chain-of-Thought (CoT) data via a multi-persona strategy. To ensure data quality, these raw CoTs undergo a self-filtering process, utilizing multi-agents to guarantee their logical rigor and robustness. Finally, the evaluator trained on the refined data is deployed as a reward model to guide the story generation task. Experimental results demonstrate that our framework achieves state-of-the-art (SOTA) performance on three evaluation benchmarks including StoryER, HANNA and OpenMEVA. Furthermore, when served as a reward model, it significantly enhances the quality of generated stories, thereby fully validating the superiority of our self-evolving approach.
\end{abstract}

\section{Introduction}
\label{sec:intro}

\begin{figure}[t!]
\centering
\includegraphics[width=0.9\linewidth]{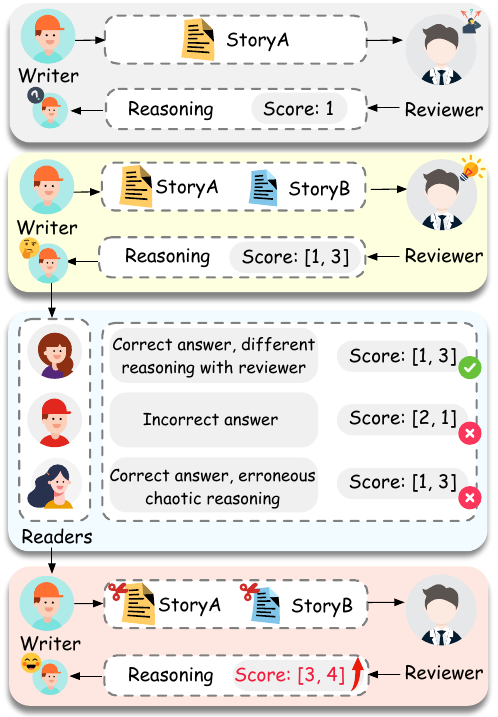}
\caption{A reviewer hesitates over scoring a single story, with feedback that feels cryptic to the writer. When assessing two stories, scores are precise, yet the feedback, though acceptable, leaves the writer unsure how to revise. The writer then turns to readers. Some reach the same conclusions but offer different suggestions, which prove helpful, and the writer crafts better stories.}
\label{fig:intro}
\vspace{-5mm}
\end{figure}

Large Language Models (LLMs) as automated evaluators, or LLMs-as-a-judge \cite{zheng2023judging,chang2024survey}, have shown immense potential across numerous tasks\cite{li2024leveraging, gao2025llm}. However, their capabilities remain limited in open-ended creative domains like story evaluation \cite{li2025generation}, which demand a deep understanding of plot, character, and creativity \cite{yang2024makes}. An accurate story evaluator is not only crucial for assisting human quality judgments but, more importantly, can serve as a vital reward signal \cite{ouyang2022training, stiennon2020learning} for automated story generation systems, guiding them to produce higher-quality narratives \cite{gomez2023confederacy}.

Current approaches to building story evaluators face a significant bottleneck. Although prompt engineering for proprietary models \cite{liu2023g,chiang2023closer,liu2023calibrating} offers flexibility, its results can be unstable and generalize poorly \cite{zhao2021calibrate}. The alternative mainstream path, fine-tuning open source models, confronts a challenge: existing fine-tuning paradigms are mostly designed for general-purpose Natural Language Generation (NLG) tasks \cite{li2023generative,hu2024llm}. Even when they incorporate story evaluation data \cite{hu2024themis,li2023generative,xu2023instructscore,jiang2023tigerscore}, they are ill-equipped for the fine-grained demands of story assessment. Moreover, even specialized story evaluators like Coke \cite{joshi2025coke}, are trained on human judgments without a step-by-step reasoning process, thereby limiting its potential as a high-fidelity judge.

The reason before predict paradigm, or Chain of Thought (CoT) \cite{wei2022chain} which can improve prediction explainability by generating reasoning steps, and it boosts predictive accuracy \cite{wei2022chain,wang2023scott}. Meanwhile, various methods have been proposed to improve a model's reasoning capabilities by enhancing the quality of CoT. These include techniques to bolster result stability during inference by having the model sample multiple different reasoning paths for the same problem \cite{wang2022self}; methods to distill the complex reasoning abilities of large models like GPT-4 onto smaller ones \cite{hsieh2023distilling, mukherjee2023orca}; and paradigms where models continuously strengthen their reasoning skills through self-learning \cite{zhou2024self,zelikman2024star}. Nevertheless, the domain of story evaluation currently lacks a methodology specifically dedicated to this CoT approach.

To overcome this critical gap, we propose the Self-Evolving Pairwise Reasoning (EvolvR) framework, designed to instill rigorous evaluation and reasoning capabilities into open-source models through self-driven data evolution. We build our framework on pairwise comparison because our analysis of human-annotated datasets reveals that this format exhibits significantly higher rating consistency than pointwise scores (see supplementary materials for details) and captures nuanced human preferences. As shown in the scenario in Figure \ref{fig:intro}, pairwise comparison and multi-perspective reasoning are conducive to story evaluation and creation. We devise a multi-role self-synthesis strategy for the model to autonomously generate a large corpus of pairwise comparison data, where each sample is augmented with a detailed, score-aligned CoT rationale. Subsequently, to ensure the logical rigor of these synthetic thoughts, we introduce a multi-agent self-filtering and evolution mechanism that purifies and enhances the data quality.

Our main contributions are as follows:
\begin{itemize}
    \item We propose a self-evolving framework grounded in pairwise comparison (EvolvR) which is featuring a novel multi-persona strategy for CoT self-synthesis and a multi-agent mechanism for self-filtering and evolution. This provides a scalable solution to the scarcity of high-quality data for complex reasoning tasks.
    \item We achieve new state-of-the-art (SOTA) performance on three authoritative story evaluation benchmarks including StoryER \cite{chen2023storyer}, HANNA \cite{chhun2024language}, and OpenMEVA \cite{guan2021openmeva}, demonstrating the superior accuracy of our evaluation model.
    \item We validate that our EvolvR-trained evaluator serves as an effective reward model, significantly improving the quality of generated stories when guiding a generation task. This confirms the practical utility and superiority of our approach.
\end{itemize}

Our code will be available at \url{https://github.com/xindaaW/EvolvR}.

\begin{figure*}[t!]
\centering
\includegraphics[width=0.95\linewidth]{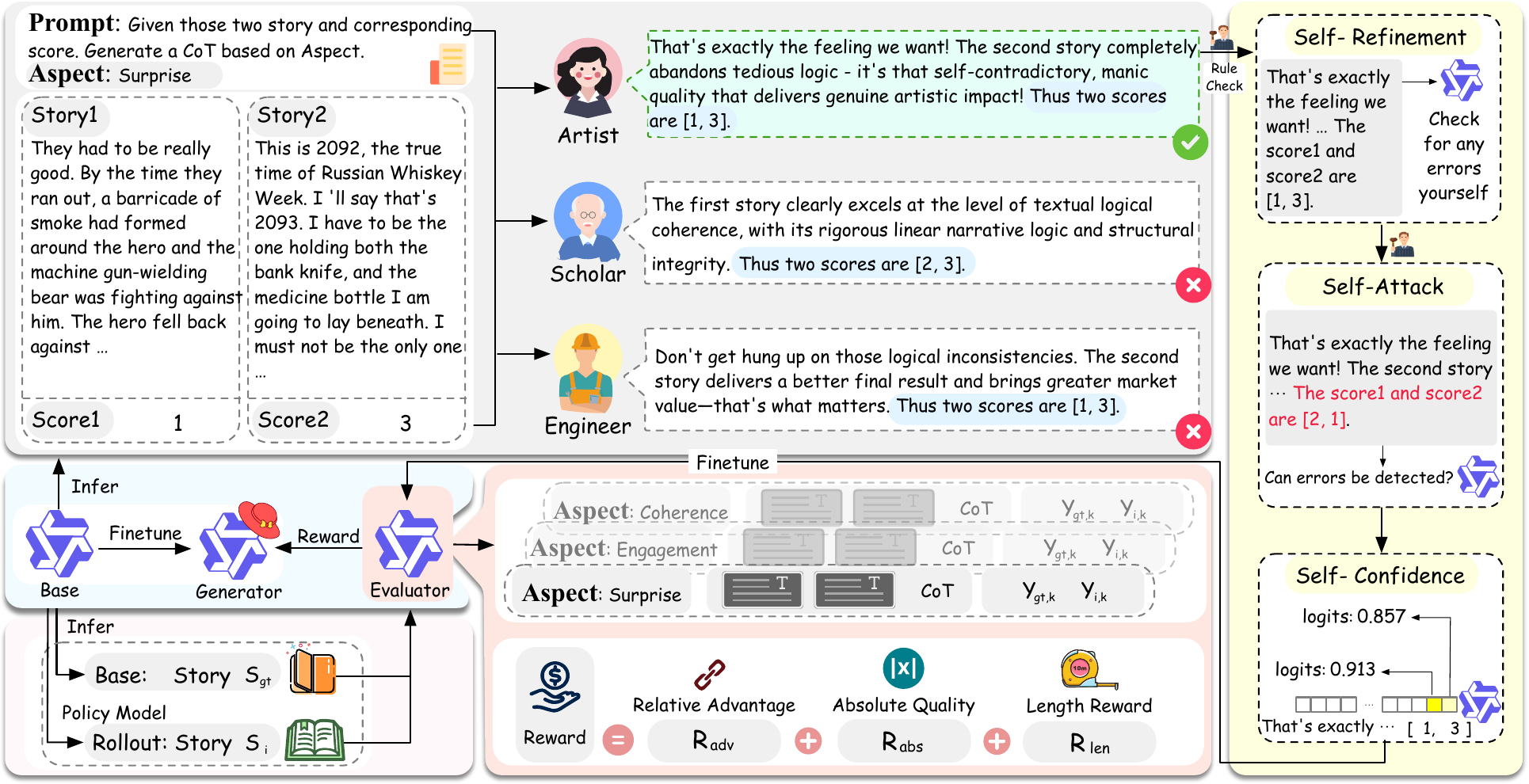}
\caption{The EvolvR Framework. We self-synthesize a diverse set of CoT rationales via a multi-persona strategy, which are refined through a multi-agent evolution pipeline to ensure high quality, and the trained evaluator is deployed as a reward model to guide and enhance story generation.}
\label{fig:method}
\vspace{-4mm}
\end{figure*}

\section{Related Work}
\label{sec:related_work}

\subsection{Story Evaluation with Large Language Models}
Large Language Models (LLMs) have driven a paradigm shift in NLG evaluation. As evaluators, LLMs show higher consistency with human judgment than traditional metrics like BLEU \cite{papineni2002bleu} and ROUGE \cite{lin2004rouge} and provide interpretable rationales, enhancing evaluation reliability \cite{wang2023chatgpt, gao2025llm}. LLM-based story evaluation has advanced along several avenues. The first is prompt-based evaluation, using powerful LLMs (e.g., GPT-4) as zero-shot or few-shot evaluators \cite{lee2024checkeval}. Quality is enhanced through strategies like task decomposition into dimensions like plot and character \cite{gong2023coascore}, or multi-agent debate frameworks such as ChatEval \cite{chan2023chateval} and SCALEEVAL \cite{chern2024can}. MATEval notably incorporates CoT and self-reflection to boost performance \cite{li2024mateval}. The second direction is training-based expert models to address the cost and reproducibility issues of API methods. This includes general-purpose evaluators like Prometheus \cite{kim2023prometheus}, Themis \cite{hu2024themis}, and TigerScore \cite{jiang2023tigerscore}, as well as specialized models for dimensions like coherence or interestingness including COHESENTIA \cite{maimon2023cohesentia} and PERSE \cite{wang2024learning}. A third approach utilizes LLM-based probabilities, where methods like GPTScore \cite{fu2023gptscore} use conditional generation probability as a quality proxy, and DELTASCORE \cite{xie2023deltascore} measures probability changes after text perturbation. Despite these advances, LLM-based methods face persistent challenges. Beyond documented issues like positional \cite{wang2023large} and knowledge bias \cite{liu2023g}, a more fundamental problem is their limited reasoning capability \cite{yang2024makes}, which can cause a logical disconnect or contradiction between the evaluation rationale and the final score, limiting their fidelity.

\subsection{Story Generation with Large Language Models}
LLMs have achieved significant breakthroughs in creating long-form, coherent, and creative narratives \cite{yang2024makes}. Research in LLM-based story generation focuses on two core challenges: controllability and long-form coherence. For controllability, LLMs can effectively adhere to control signals ranging from high-level themes to fine-grained character details \cite{guan2022lot, fan2018hierarchical, vijjini2022towards}. For long-form coherence, the hierarchical plan-and-write framework is the predominant approach \cite{you2023eipe}. This paradigm, which first generates an outline and then the full text, is crucial for maintaining global consistency and has evolved from basic frameworks to advanced models using tree-structured outlines for fine-grained control \cite{yao2019plan, yang2022doc}. To enhance quality beyond one-shot generation, researchers have also explored iterative frameworks like recursive revision \cite{yang2022re3} and interactive systems with dynamic memory \cite{zhou2023recurrentgpt}. Despite success in structural control, existing methods cannot directly optimize for holistic story quality. RLHF is the ideal paradigm to address this \cite{ouyang2022training}, but has been hindered by the absence of an accurate reward model for narrative quality. To our knowledge, our work is the first to overcome this bottleneck. By developing a high-fidelity evaluator, we unlock RLHF for story generation, enabling direct enhancement of narrative artistry.

\section{Method}
\label{sec:method}

\subsubsection{Problem Formulation}
\label{sec:problem}
Let $\mathcal{S}$ be the space of stories. Our goal is to train an evaluator model $R_\phi$ parameterized by $\phi$. Given a story pair $(S_a, S_b) \in \mathcal{S} \times \mathcal{S}$ and a specific evaluation aspect $k$ from a set of $K$ aspects (e.g., Creativity, Surprise), the model should produce a pair of scores $(y_{a,k}, y_{b,k})$, where $y \in [1, 5]$. 

Our methodology is designed to create a high-fidelity story evaluator by training it on systematically generated and filtered reasoning paths. This evaluator then serves as a robust reward model to enhance story generation through reinforcement learning. As shown in Figure \ref{fig:method}, the process unfolds in three main stages: (1) self-synthesis of score-aligned CoTs, (2) a multistage CoT evolution and selection pipeline, and (3) story generation with the evaluator.

\subsection{Self-Synthesis of Score-Aligned CoTs}
\label{sec:synthesis}

The core of our approach is to leverage CoT reasoning. We define the evaluator's task as learning a mapping:
\begin{equation}
R_\phi(S_a, S_b, k) = (C_k;y_{a,k};y_{b,k})
\end{equation}
where $C_k$ is a textual rationale that logically justifies the assigned scores $(y_{a,k}, y_{b,k})$ for aspect $k$. To achieve this, we first need a dataset of high-quality tuples $\mathcal{T} = \{(S_a, S_b, k, C_k, y_{a,k}, y_{b,k})\}_i$.

\subsubsection{Multi-Persona CoT Synthesis}
As such data is not readily available, we synthesize it. We start with a seed dataset of story pairs with ground-truth scores, $\{(S_a, S_b, k, y_{a,k}^*, y_{b,k}^*)\}$. For each tuple, we prompt a LLM, which we denote as $LLM_{\text{self}}$, to generate a candidate derivation. 

To foster diversity, we employ a multi-persona approach. We define a set of personas $\mathcal{P} = \{p_1, \dots, p_m\}$, where each persona represents a distinct viewpoint (e.g. academic personality, artistic personality, sharp-tongued personality and so on). For each sample and each persona $p_m \in \mathcal{P}$, we generate a candidate derivation:
\begin{equation}
(C_{k}^m;y_{a,k};y_{b,k}) \sim LLM_{\text{self}}(\cdot | S_a, S_b, k, y_{a,k}^*, y_{b,k}^*, p_m)
\end{equation}
This process yields a large, diverse initial pool of candidate derivation, $\mathcal{D}_{\text{pool}}$, for each data point. The detailed prompts are provided in supplementary materials.

\subsection{CoTs Evolution and Selection Pipeline}
\label{sec:selection}

The raw derivations in $\mathcal{D}_{\text{pool}}$ are noisy. We introduce a rigorous, multi-agents pipeline to filter and refine them, ensuring logical consistency, robustness, and alignment with the target scores. This pipeline, detailed in Algorithm~\ref{alg:cot_pipeline}, applies a sequence of four evolutionary selection operators.

\paragraph{Self-Rulecheck ($\mathcal{F}_{\text{rule}}$)}
This filter is applied to each candidate derivation $D_i$ from a given pool $\mathcal{D}_{\text{pool}}$. It ensures that the final conclusion explicitly stated within $D_i$ is consistent with the ground truth scores $(y_a^*, y_b^*)$. We employ a deterministic parsing function, $\mathrm{ParseScores}(\cdot)$, to extract the final scores $(y'_a, y'_b)$ from the text of $D_i$. The CoT passes the check if the extracted scores exactly match the ground truth scores.
\begin{equation}
\mathcal{F}_{\text{rule}}(C_{i,k}^m;y'_{a,k};y'_{b,k}) = \mathbb{I}\left( (y'_{a,k}, y'_{b,k}) = (y_{a,k}^*, y_{b,k}^*) \right)
\end{equation}
where $(y'_{a,k}, y'_{b,k}) = \mathrm{ParseScores}(C_{i,k}^m;y'_{a,k};y'_{b,k})$, and $\mathbb{I}(\cdot)$ is the indicator function.

\paragraph{Self-Refinement ($\mathcal{F}_{\text{refine}}$)}
This operator leverages the model's intrinsic ability for self-improvement. Applied to each candidate  derivation $D_i$ from a given pool $\mathcal{D}_{\text{pool}}$, the model itself ($LLM_{\text{self}}$), guided by a refinement agent $A_{\text{refine}}$, improves the logical flow and clarity of the rationale. This process generates a refined version $C'_{i,k}$ from the text of $D_i$ while preserving the original reasoning.
\begin{equation}
\mathcal{F}_{\text{refine}}(C_{i,k}^m) = LLM_{\text{self}}(\cdot | A_{\text{refine}}, C_{i,k}^m)
\end{equation}
where $A_{\text{refine}}$ represents the agent for the self-refinement task. For subsequent steps, we denote the refined candidates simply as $C_{i,k}^m$.

\paragraph{Self-Attack ($\mathcal{F}_{\text{attack}}$)}
We assess its logical robustness by creating a corrupted version, $(C_{i,k}^m;y_{a,k}^{corr};y_{b,k}^{corr})$, where the final scores are replaced to contradict the rationale. The same model ($LLM_{\text{self}}$) is then prompted to act as a judge, guided by agent $A_{\text{attack}}$, to check for inconsistencies. The original CoT $C_{i,k}^m$ is considered robust only if the model successfully identifies the contradiction in its own altered reasoning.
\begin{equation}
\begin{split}
\mathcal{F}_{\text{attack}}&(C_{i,k}^m; y_{a,k}^{corr}, y_{b,k}^{corr}) \\
    &= \mathbb{I}\Big(\mathrm{DetectsContradiction}\big(LLM_{\text{self}}(A_{\text{attack}}, \\
    &\quad (C_{i,k}^m; y_{a,k}^{corr}; y_{b,k}^{corr}))\big)\Big)
\end{split}
\end{equation}
where $A_{\text{attack}}$ represents the agent for the contradiction detection task.

\paragraph{Self-Confidence ($\mathcal{F}_{\text{confidence}}$)}
It selects for CoTs that lead the model to predict the ground truth scores with high confidence. We inspect the output logits of the model. A CoT $C_{i,k}^m$ passes if the logit corresponding to the ground truth score token is the maximum for each story.
\begin{equation}
\begin{split}
\mathcal{F}_{\text{confidence}}&(C_{i,k}^m, y_{a,k}^*, y_{b,k}^*) \\
    &= \mathbb{I}\left( \operatorname*{argmax}_{y} L(y|S_a, C_{i,k}^m) = y_{a,k}^* \right) \\
    &\quad \land \mathbb{I}\left( \operatorname*{argmax}_{y} L(y|S_b, C_{i,k}^m) = y_{b,k}^* \right)
\end{split}
\end{equation}
where $L(y|S_a, C_{i,k}^m)$ represents the logit from the model.

\begin{algorithm}[tb]
\caption{Pipeline of CoTs Evolution and Selection}
\label{alg:cot_pipeline}
\textbf{Require}: $\mathcal{D}_{\text{pool}}$: The initial pool of candidate derivations $\{D_i\}$. Each $D_i$ is a tuple $(C_{i,k}^m, y_{a,k}, y_{b,k})$. $y_{a,k}^*, y_{b,k}^*$: The ground-truth scores for aspect $k$. \\
\textbf{Ensure}: $\mathcal{D}_{\text{final}}$: The final pool of high-quality and evolved CoTs and scores.
\begin{algorithmic}[1]
\STATE $\mathcal{P} \gets [\mathcal{F}_{\text{rule}}, \mathcal{F}_{\text{refine}}, \mathcal{F}_{\text{rule}}, \mathcal{F}_{\text{attack}}, \mathcal{F}_{\text{confidence}}]$
\STATE $\mathcal{D}_{\text{final}} \gets \emptyset$
\FOR{each candidate derivation $D_i \in \mathcal{D}_{\text{initial}}$}
    \STATE $D_{\text{current}} \gets D_i$
    \STATE $\text{survived} \gets \text{true}$ 
    \FOR{each operator $\mathcal{F}$ in pipeline $\mathcal{P}$}
        \IF{$\mathcal{F}$ is a refinement operator}
            \STATE $D_{\text{current}} \gets \mathcal{F}(D_{\text{current}})$ 
        \ELSE 
            \IF{$\mathcal{F}(D_{\text{current}}, y_a^*, y_b^*) = \text{false}$}
                \STATE $\text{survived} \gets \text{false}$
                \STATE \textbf{break}
            \ENDIF
        \ENDIF
    \ENDFOR
    \IF{survived}
        \STATE $\mathcal{D}_{\text{final}} \gets \mathcal{D}_{\text{final}} \cup \{D_{\text{current}}\}$
    \ENDIF
\ENDFOR
\STATE \textbf{return} $\mathcal{D}_{\text{final}}$  
\end{algorithmic}
\end{algorithm}

\begin{table*}[t!]
\centering
\renewcommand{\arraystretch}{1.1} 
\setlength{\tabcolsep}{1.5mm}
\scalebox{0.9}{
\begin{tabular}{@{}l *{10}{c}@{}} 
\toprule
& \multicolumn{5}{c}{\textbf{StoryER Dataset}} & \multicolumn{5}{c}{\textbf{HANNA Dataset}} \\
\cmidrule(lr){2-6} \cmidrule(l){7-11}
\textbf{Model} & Pearson  & Spearman  & Kendall  & MSE  & F1-Score  & Pearson  & Spearman  & Kendall  & MSE  & F1-Score  \\
\midrule
\multicolumn{11}{l}{\textit{General Closed-source Models}} \\
GPT-4o                      & 0.4808 & 0.4679 & 0.3975 & \underline{0.0398} & 0.8979 & 0.4270 & 0.3517 & 0.2867 & 0.0744 & 0.4922 \\
GPT-4.1                     & 0.3673 & 0.2908 & 0.2567 & 0.0705 & 0.8998 & 0.3171 & 0.2987 & 0.2612 & 0.0820 & 0.6727 \\
GPT-o3-0416-global          & \underline{0.5682} & 0.4944 & 0.4402 & 0.0477 & 0.8703 & 0.4837 & 0.4128 & 0.3543 & 0.0854 & 0.4016 \\
Gemini-2.5-flash-06-17        & 0.3882 & 0.3042 & 0.2701 & 0.1091 & 0.9173 & 0.3883 & 0.3647 & 0.3111 & 0.1876 & 0.5700 \\
Gemini-2.5-Pro-06-17        & 0.5451 & \underline{0.5275} & \underline{0.4796} & 0.0931 & 0.8866 & 0.5000 & 0.4586 & 0.4041 & 0.1782 & 0.4569 \\
Claude-4-Sonnet             & 0.4157 & 0.3415 & 0.3024 & 0.0676 & 0.8662 & 0.4053 & 0.3675 & 0.3273 & 0.1327 & 0.3659 \\
Claude-4-opus4             & 0.4796 & 0.4060 & 0.3561 & 0.0995 & 0.8440 & 0.4872 & 0.4237 & 0.3702 & 0.1814 & 0.3833 \\
\midrule
\multicolumn{11}{l}{\textit{NLG Evalution Open-Source Models}} \\
InstructScore-7B \cite{xu2023instructscore}            & 0.1902 & 0.1664 & 0.1251 & 0.1303 & 0.8430 & 0.1258 & 0.1056 & 0.0989 & 0.2071 & 0.5787 \\
Themis-8B \cite{hu2024themis}                   & 0.5362 & 0.3870 & 0.3484 & 0.0743 & 0.9459 & 0.4565 & 0.3696 & 0.3295 & 0.1432 & 0.3799 \\
TIGERScore-13B \cite{jiang2023tigerscore}              & 0.2759 & 0.2364 & 0.2317 & 0.1455 & 0.8637 & 0.2897 & 0.2056 & 0.1879 & 0.1309 & 0.5180 \\
AutoJ-13B \cite{li2023generative}               & 0.3182 & 0.1873 & 0.1631 & 0.0623 & 0.9012 & 0.4153 & 0.3689 & 0.3223 & 0.0609 & 0.4410 \\
\midrule
\multicolumn{11}{l}{\textit{Story Evalution Finetune Open-source Models}} \\
Coke                        & 0.3142 & --- & --- & 0.0812 & 0.6509 & --- & --- & --- & --- & --- \\
Qwen2.5-7B-Instruct         & 0.2206 & 0.1611 & 0.1381 & 0.1434 & 0.8308 & 0.1873 & 0.1605 & 0.1339 & 0.1499 & 0.5236 \\
+ Point CoT                 & 0.5677 & 0.4784 & 0.4392 & \textbf{0.0293} & 0.9344 & 0.5173 & 0.4919 & 0.4380 & 0.0521 & 0.6715 \\
+ GRPO                      & 0.3430 & 0.2436 & 0.2178 & 0.0849 & 0.8793 & 0.1773 & 0.1717 & 0.1518 & 0.0790 & 0.4507 \\
+ Point CoT + GRPO          & 0.5646 & 0.4831 & 0.4472 & 0.0419 & \underline{0.9433} & \underline{0.5307} & \underline{0.5121} & \underline{0.4606} & \underline{0.0483} & \underline{0.6829} \\
\textbf{EvolvR} & \textbf{0.6774}  & \textbf{0.6000}  & \textbf{0.5353}  & 0.0528  & \textbf{0.9474}  & \textbf{0.6155} & \textbf{0.6033} & \textbf{0.5429} & \textbf{0.0440} & \textbf{0.7406} \\
\bottomrule
\end{tabular}
}
\caption{Performance comparison on the StoryER and HANNA datasets. EvolvR, is compared against baselines and analyzed via an ablation study across both benchmarks. We report Pearson correlation ($\uparrow$), Spearman correlation ($\uparrow$), Kendall correlation ($\uparrow$), Mean Squared Error (MSE $\downarrow$), and F1-Score ($\uparrow$). Values in parentheses indicate improvement over the Base Model.}
\label{tab:main_result}
\vspace{-3mm}
\end{table*}

\subsection{Story Generation with the Evaluator}
\label{sec:generation}

Our trained evaluator, $R_\phi$, provides a high-fidelity reward signal for fine-tuning a story generation policy, $\pi_\theta$, using the Group Relative Policy Optimization (GRPO) algorithm\cite{shao2024deepseekmath}.

Following the GRPO framework, for each prompt $q$, we first sample a group of $G$ candidate stories $\{S_1, \dots, S_G\}$ from the current policy $\pi_\theta$. Next, we compute a reward $r_i$ for each candidate $S_i$ using our custom reward function. This function evaluates $S_i$ against a common, high-quality reference story $S_{\text{gt}}$, which is pre-generated from the base model to serve as a stable anchor. The reward $r_i = \mathcal{R}(S_i, S_{\text{gt}})$ is a weighted composite of three key components:
\begin{equation}
\mathcal{R}(S_i, S_{\text{gt}}) = \mathcal{R}_{\text{adv}} + \mathcal{R}_{\text{abs}} + \mathcal{R}_{\text{len}}
\end{equation}
where the components include relative advantage, absolute quality, and length reward, which are defined as:
\begin{align}
    \mathcal{R}_{\text{adv}} &= w_1 \cdot \sum_{k=1}^{K} \alpha_k (y_{i,k} - y_{\text{gt},k}) \\
    \mathcal{R}_{\text{abs}} &= w_2 \cdot \sum_{k=1}^{K} \alpha_k y_{i,k} \\
    \mathcal{R}_{\text{len}} &= w_3 \cdot f(\text{len}(S_i))
\end{align}
Here, $y_{i,k}$ and $y_{\text{gt},k}$ are the scores assigned by our evaluator $R_\phi$, $\alpha_k$ is a weight controlling the importance of the $k$-th evaluation aspect, and $w_j$ are weighting hyperparameters.

The advantage for each candidate sequence $o_i$ is then calculated by comparing its reward to the group's average:
\begin{equation}
    \label{eq:advantage}
    \hat{A}_i = r_i - \frac{1}{G}\sum_{j=1}^{G} r_j
\end{equation}
This sequence-level advantage is applied at each token step (i.e., $\hat{A}_{i,t} = \hat{A}_i$ for all $t$) and used to update the policy by maximizing the GRPO objective:
\begin{align}
\label{eq:grpo_objective_manual_align}
\mathcal{J}_{\text{GRPO}}(\theta) &= \mathbb{E}_{q \sim P(Q), \{o_i\} \sim \pi_{\theta_{\text{old}}}}\bigg[ \frac{1}{G} \sum_{i=1}^{G} \frac{1}{|o_i|} \sum_{t=1}^{|o_i|} \Big\{ \nonumber \\
& \hspace{-2em} \min\Big( \rho_{i,t}(\theta) \hat{A}_{i,t}, \text{clip}(\rho_{i,t}(\theta), 1-\epsilon, 1+\epsilon)\hat{A}_{i,t} \Big) \nonumber \\
& \hspace{-2em} - \beta D_{\text{KL}}(\pi_\theta || \pi_{\text{ref}}) \Big\} \bigg]
\end{align}
where the token-level policy ratio is $\rho_{i,t}(\theta) = \frac{\pi_{\theta}(o_{i,t}|q, o_{i,<t})}{\pi_{\theta_{\text{old}}}(o_{i,t}|q, o_{i,<t})}$, $\epsilon$ is the clipping parameter, and $\beta$ controls the KL penalty.




\section{Experiments}
\subsection{Experimental Settings}
\subsubsection{Datasets}
Our evaluation is conducted on two public benchmarks with human annotations: StoryER \cite{chen2023storyer} and HANNA \cite{chhun2024language}. Both contain stories from WritingPrompts \cite{kroll1994guidelines} with multi-dimensional quality ratings, and we follow their official train/test splits to ensure reproducibility. To test generalization, we also report zero-shot performance on the OpenMEVA benchmark \cite{guan2021openmeva}. Details of these datasets can be found in the supplementary material. In our generation setup, the evaluator, trained on the HANNA dataset, functions as a reward model to steer a generator prompted with inputs from the same dataset.

\subsubsection{Baselines}
We compare our model against three categories of baselines. First, we evaluate proprietary large language models via prompt engineering. This category includes models from OpenAI (GPT-4o, GPT-4.1, and GPT-o3-0416-global \cite{hurst2024gpt}), Google (Gemini-2.5-flash and Gemini-2.5-Pro \cite{comanici2025gemini}), and Anthropic (Claude-4-Sonnet and Claude-4-opus4). Second, we compare against open-source evaluators fine-tuned on various NLG tasks, such as summarization, machine translation, and story generation. This group includes TIGERScore \cite{jiang2023tigerscore}, InstructScore \cite{xu2023instructscore}, Themis \cite{hu2024themis}, and AutoJ \cite{li2023generative}. Notably, models like TIGERScore and InstructScore require a ground-truth reference text, a constraint our method does not have. Finally, we conduct comparisons with methodologically similar models. This includes Coke \cite{joshi2025coke}, which is also trained on the StoryER dataset but requires human-written comments as input. To analyze the contribution of our design choices, we also establish our own baselines: (1) the base Qwen2.5-7B-Instruct model \cite{team2024qwen2} without any fine-tuning, and (2) ablated versions of our model with different components of our proposed framework removed.




\subsubsection{Evaluation Metrics}
Following established practices in NLG evaluation, we assess the performance of all models by measuring the agreement between their predicted ratings and the ground-truth human judgments. Specifically, we calculate three standard correlation coefficients: Pearson's correlation coefficient ($r$), Spearman's rank correlation coefficient ($\rho$), Kendall's rank correlation coefficient ($\tau$), In addition, to align with the evaluation protocol of prior work like Coke \cite{joshi2025coke}, we also report the Mean Squared Error (MSE) on normalized scores and the F1-Score, providing a comprehensive view of both correlation and accuracy.

To evaluate the quality of the stories produced by our fine-tuned generative model, we employed a comprehensive human evaluation protocol. We enlisted a panel of four professional screenwriters as expert judges to ensure nuanced and domain-specific assessment. These judges were asked to rate the generated stories on a 1-to-5 scale across six key dimensions, mirroring the criteria from the HANNA dataset: Relevance, Coherence, Empathy, Surprise, Engagement, and Complexity. Based on their detailed assessments, we report two primary metrics. The first is the Average Score, calculated by averaging the scores across all six dimensions for each model, where a higher score indicates superior quality. The second metric is the Win Rate, where we perform head-to-head comparisons of our model's generated stories against two distinct references: (1) stories generated by the base model without any fine-tuning and ((2) stories from the original HANNA dataset.

\begin{table}[t!]
\centering
\renewcommand{\arraystretch}{1.1} 
\setlength{\tabcolsep}{1.2mm}
\scalebox{0.9}{
\begin{tabular}{@{}lccc@{}} 
\toprule
\textbf{Model} & Pearson & Spearman & Kendall \\
\midrule
\multicolumn{4}{l}{\textit{NLG Evaluation Open-Source Models}} \\
InstructScore-7B\cite{xu2023instructscore}             & 0.1578 & 0.1257 & 0.1046 \\
Themis-8B\cite{hu2024themis}                   & 0.1469 & 0.1719 & 0.1457 \\
TIGERScore-13B\cite{jiang2023tigerscore}                 & 0.1661 & 0.1649 & 0.1541 \\
AutoJ-13B\cite{li2023generative}                    & 0.1871 & 0.1952 & 0.1611 \\
\midrule
\multicolumn{4}{l}{\textit{Finetune Open-source Models}} \\
Qwen2.5-7B-Instruct           & 0.1061 & 0.1175 & 0.0976 \\
+ Pointwise CoT               & 0.1474 & 0.1503 & 0.1291 \\
+ GRPO                        & 0.0947 & 0.0994 & 0.0897 \\
+ Pointwise CoT + GRPO        & 0.1605 & 0.1654 & 0.1415 \\
\textbf{EvolvR}         & \textbf{0.2092} & \textbf{0.2087} & \textbf{0.1881} \\
\bottomrule
\end{tabular}
}
\caption{Black-box evaluation results on the OpenMEVA benchmark. Our method is compared against existing open-source story evaluation models. All metrics are correlation coefficients, where higher is better (↑).}
\label{tab:openmeva_results}
\vspace{-3mm}
\end{table}

\subsection{Main Results and Analysis}
\subsubsection{Analysis of Evaluation Results}
The main results, presented in Table \ref{tab:main_result}, demonstrate that our proposed model, EvolvR, achieves SOTA performance in correlation with human judgments on both datasets, while also showing strong results on MSE and F1-Score. 

\textit{a) Comparison with Proprietary LLMs: }
In the GPT series, the deep reasoning model GPT-o3 outperforms GPT-4o and GPT-4.1. Similarly, Gemini-Pro surpasses its lightweight counterpart, Gemini-flash, and Claude-opus4 is superior to the faster Claude-Sonnet. This trend strongly suggests that a more profound thought process is crucial for complex tasks like story evaluation. This observation directly supports us that a high-quality evaluation stems from a deep and rigorous CoT, which our framework is explicitly designed to generate and refine.

\begin{table}[t!]
\centering
\renewcommand{\arraystretch}{1.1} 
\setlength{\tabcolsep}{1.5mm}
\scalebox{0.9}{
\begin{tabular}{@{}lccc@{}} 
\toprule
\textbf{Model} & Pearson ↑ & Spearman ↑ & Kendall ↑\\
\midrule
Qwen3-4B  & 0.1322 & 0.1606 & 0.1361 \\
+ GRPO       & 0.1374 & 0.1596 & 0.1357 \\
+ Pointwise CoT  & 0.4998 & 0.4804 & 0.4255 \\
+ Pairwise CoT   & \textbf{0.5855} & \textbf{0.5755} & \textbf{0.5214} \\
\midrule
LLama3.1-8B  & 0.1490 & 0.1682 & 0.1034 \\
+ GRPO       & 0.2288 & 0.3197 & 0.2710 \\
+ Pointwise CoT  & 0.5244 & 0.5092 & 0.4538 \\
+ Pairwise CoT   & \textbf{0.5852} & \textbf{0.5826} & \textbf{0.5291} \\
\midrule
Qwen3-8B  & 0.2236 & 0.2045 & 0.1714 \\
+ GRPO       & 0.2855 & 0.2289 & 0.1888 \\
+ Pointwise CoT  & 0.5302 & 0.5192 & 0.4611 \\
+ Pairwise CoT   & \textbf{0.6092} & \textbf{0.5920} & \textbf{0.5334} \\
\midrule
Qwen3-14B     & 0.2676 & 0.2646 & 0.2232 \\
+ GRPO       & 0.2728 & 0.2688 & 0.2204 \\
+ Pointwise CoT  & 0.5049 & 0.4747 & 0.4232 \\
+ Pairwise CoT   & \textbf{0.6421} & \textbf{0.6349} & \textbf{0.5753} \\
\bottomrule
\end{tabular}
}
\caption{The performance of Pairwise CoT, Pointwise CoT Fine-tuning, and GRPO Reinforcement Learning experiments on the HANNA dataset across a diverse set of base models with varying scales and architectures (including the Qwen3 and LLaMA3.1 series).}
\label{tab:sft_vs_evolvr_comparison}
\vspace{-3mm}
\end{table}

\begin{table*}[t]
\centering
\renewcommand{\arraystretch}{1.1}
\scalebox{0.9}{
\begin{tabular}{@{}lcccccc@{}}
\toprule
\textbf{Model} & \textbf{Relevance} & \textbf{Coherence} & \textbf{Empathy} & \textbf{Surprise} & \textbf{Engagement} & \textbf{Complexity} \\
\midrule
Qwen2.5-7B-Instruct     & 4.4000 & \textbf{4.3375} & 2.6375 & 2.7375 & \underline{3.1000} & 3.5750 \\
\midrule
+ SFT                   & 3.4875 (-20.7\%) & 3.6125 (-16.7\%) & 2.7000 (+2.4\%) & 2.8250 (+3.2\%) & \underline{3.1000} (0.0\%) & 3.2875 (-8.0\%) \\
+ Point-RM GRPO         & \underline{4.4875} (+2.0\%) & \underline{4.2625} (-1.7\%) & \textbf{2.8375} (+7.6\%) & \underline{2.9500} (+7.8\%) & 3.0875 (-0.4\%) & \underline{3.6250} (+1.4\%) \\
\textbf{EvolvR GRPO}    & \textbf{4.5875} (+4.3\%) & 4.2500 (-2.0\%) & \underline{2.7625} (+4.7\%) & \textbf{2.9625} (+8.2\%) & \textbf{3.2500} (+4.8\%) & \textbf{3.7250} (+4.2\%) \\
\bottomrule
\end{tabular}
}
\caption{Multi-dimensional average scores. Values in parentheses denote the percentage change relative to the baseline.}
\label{tab:dimension_scores}
\vspace{-4mm}
\end{table*}

\textit{b) Comparison with Open-Source NLG Evaluators: }
Our model significantly surpasses general-purpose NLG evaluators like InstructScore, TIGERScore, Themis, and AutoJ, even though they have been trained on story-related data. First, models like InstructScore and TIGERScore are constrained by their need for a reference answer, a requirement that is often impractical in real-world creative scenarios where no single correct story exists. Second, the performance gap with Themis and AutoJ which operate without references validates the superiority of our framework's approach: to self-synthesize and evolutionarily select high-quality CoTs tailored specifically for story evaluation. To validate the effectiveness of our model in broader scenarios, we compared it against several mainstream open-source NLG evaluation models on the OpenMEVA benchmark, where the performance of all models was observed to be lower due to the dataset's use of a single holistic score instead of dimension-specific ratings. The detailed results are shown in Table~\ref{tab:openmeva_results}.

\textit{c) Comparison with Specialized Story Evaluators: }
The comparison with Coke, a model also trained on StoryER, is particularly revealing. Coke's methodology relies on processing human-written comments by extracting keywords directly for scoring. Our model's performance demonstrates that proactively constructing and evolving high-fidelity rationales from scratch is a more robust and effective strategy than passively relying on existing, often noisy and inconsistent, human commentary.

\subsubsection{Internal Methodological Comparison}
Finally, our internal results show that EvolvR is the primary driver of performance. It is noteworthy that the subsequent application of GRPO does not yield a similarly dramatic improvement in the evaluation metrics. To further dissect the relative contributions of the key components in our framework including Pairwise CoT, Point CoT, and GRPO. We conducted a series of controlled experiments on the HANNA dataset across a diverse set of base models with varying scales and architectures including the Qwen3 \cite{yang2025qwen3} and LLaMA3.1 \cite{grattafiori2024llama} series. The results are presented in Table~\ref{tab:sft_vs_evolvr_comparison}. A universal trend across all tested models is the dramatic, qualitative leap in performance upon the introduction of fine-tuning with our self-synthesized and evolved CoT data. The results consistently show that, across all models, training with pairwise CoT significantly outperforms pointwise CoT, as the former cultivates a more discerning evaluation capability by forcing the model to focus on the fine-grained distinctions that make one story superior to another. A noteworthy phenomenon is that applying GRPO directly to the base models yields limited performance gains, and in some cases, almost no improvement at all. Further analysis reveals this is due to reward hacking and experiment details can be seen in supplementary material.

\begin{table}[t!]
\centering
\renewcommand{\arraystretch}{1.1} 
\setlength{\tabcolsep}{1.5mm}
\scalebox{0.9}{
\begin{tabular}{@{}lccc@{}} 
\toprule
\textbf{Method} & Win Base & Win HANNA & Average Score \\ 
\midrule
Qwen2.5-7B-Instruct        & ---       & 0.2249    & 3.464 \\
+ SFT                       & 0.3215    & 0.1767    & 3.168\\
+ Point-RM GRPO             & 0.5611    & 0.2909    & 3.541 \\ 
\textbf{EvolvR GRPO}         & \textbf{0.6436} & \textbf{0.3162} & \textbf{3.589} \\
\bottomrule
\end{tabular}
}
\caption{Effectiveness of EvolvR as a reward model for story generation. Win Base: the win rate against the untuned base model; Win HANNA: the win rate against human stories from the dataset; and Average Score: the mean score across all dimensions.}
\label{tab:generation_results}
\vspace{-3mm}
\end{table}


\subsubsection{EvolvR-Guided Story Generation}
We evaluate the efficacy of EvolvR as a reward model by guiding a Qwen2.5-7B-Instruct generator and comparing its performance against two baseline: SFT and GRPO guided by a standard pointwise reward model. As shown in Table~\ref{tab:generation_results}, SFT exhibits limitations in this open-ended task, learning restrictive patterns from the training data, which results in a degradation of generation quality. And the optimization via GRPO yields an overall improvement over SFT. While both the Point-RM and EvolvR models experience a slight decrease in coherence, their performance in other dimensions shows varying levels of improvement. Specifically, the EvolvR-guided model achieves the highest scores in key dimensions such as Complexity, Surprise, Relevance and Engagement (Table~\ref{tab:dimension_scores}). This suggests that a more accurate evaluator provides a more effective reward signal, enabling the generator to achieve higher-quality outputs. More details are available in the supplementary material.

\begin{table}[t!]
\centering
\renewcommand{\arraystretch}{1.1}
\setlength{\tabcolsep}{1.5mm}
\scalebox{0.9}{
\begin{tabular}{@{}lccc@{}}
\toprule
\textbf{Evolution Configuration} & Pearson ↑ & Spearman ↑ & Kendall ↑\\ 
\midrule
Baseline               & 0.5682 & 0.5591 & 0.5007 \\ 
+ Multi-persona        & 0.5941 & 0.5838 & 0.5273 \\
+ Rule-Check           & 0.5816 & 0.5792 & 0.5146 \\
+ Refinement Agent     & 0.5839 & 0.5791 & 0.5154 \\
+ Attack Agent        & 0.5989 & 0.5891 & 0.5308 \\
+ Confidence Agent     & 0.5807 & 0.5676 & 0.5100 \\
\textbf{EvolvR}  & \textbf{0.6155} & \textbf{0.6033} & \textbf{0.5429} \\ 
\bottomrule
\end{tabular}
}
\caption{Impact of each component in the EvolvR framework. This incremental study demonstrates that adding each agent module to improve performance. The results of the full EvolvR model demonstrate its optimal performance.}
\label{tab:ablation_evolution}
\vspace{-3mm}
\end{table}

\subsubsection{Impact of CoT Evolution and Selection Agent}
To validate our design, we performed an incremental analysis of the EvolvR framework, with results shown in Table~\ref{tab:ablation_evolution}. Starting from a baseline without any agents, we find that adding any single agent module leads to a clear improvement in correlation with human judgments. For example, the Multi-persona and Attack Agent modules both provide substantial gains. Ultimately, the full EvolvR model, which combines the strengths of all agents, significantly outperforms both the baseline and any single-agent variant. This confirms the effectiveness of each individual component and demonstrates the powerful synergistic effect of our complete framework design.

\section{Conclusions}
In this paper, we addressed the critical challenge of developing high-fidelity evaluators for the nuanced task of story assessment, a bottleneck for both reliable evaluation and advanced story generation. We introduced EvolvR, a novel self-evolving framework that leverages pairwise comparison to equip open-source models with deep reasoning capabilities. EvolvR autonomously generates a diverse corpus of score-aligned CoT rationales through a multi-persona strategy and subsequently refines this data via a rigorous multi-agent evolution pipeline, ensuring logical consistency and robustness. Our extensive experiments demonstrate that EvolvR achieves state-of-the-art performance on three distinct benchmarks including StoryER, HANNA, and OpenMEVA, significantly outperforming both proprietary LLMs and existing open-source evaluators. Furthermore, we validated the practical utility of our evaluator by deploying it as a reward model. The EvolvR-guided generator produced stories of higher quality, achieving superior win rates and average scores in human evaluations, thus closing the loop between evaluation and generation.
\bibliography{mainRef}
\appendix
\section{Dataset Details}

Our work leverages three prominent datasets for story evaluation: StoryER \cite{chen2023storyer}, HANNA \cite{chhun2024language}, and OpenMEVA \cite{guan2021openmeva}. We show the details in Table \ref{tab:dataset_summary}. Each dataset offers unique characteristics in terms of evaluation format, annotation dimensions, and scale, providing a comprehensive basis for training our model and assessing its performance and generalization capabilities.

\begin{table}[h!]
\renewcommand{\arraystretch}{1.2} 
\setlength{\tabcolsep}{4pt}      
\centering
\scalebox{0.85}{
\centering
\begin{tabular}{lccc}
\toprule
\textbf{Dataset} & \textbf{Samples} & \textbf{Dimensions} & \textbf{Format} \\
\midrule
StoryER-Scale   & 45,948    & 5 & Pointwise (Multi-Aspect) \\
HANNA           & 19,008    & 6 & Pointwise (Multi-Aspect) \\
OpenMEVA        & 2,000     & 1 & Pointwise (Holistic) \\
\bottomrule
\end{tabular}
}
\caption{Summary of the dataset used in our study. Each dataset provides a different combination of data scale, dimensional granularity, and evaluation format.}
\label{tab:dataset_summary}
\end{table}

\subsection{StoryER Dataset}
The StoryER dataset \cite{chen2023storyer} is a comprehensive resource for explainable story evaluation. For our research, we specifically utilize the StoryER-Scale subset. Its defining feature is that it provides not only multi-aspect scores but also human-written reasoning behind those scores. This aspect, while not directly used for training as our EvolvR framework self-synthesizes its own rationales, provides a strong motivation for our CoT-based approach and serves as a valuable reference. StoryER-Scale provides pointwise ratings on a 1-to-5 Likert scale for individual stories across multiple fine-grained dimensions. The key dimensions covered in this dataset include Coherence, Ending, Style, Character Development, and Empathy. We use StoryER-Scale as a primary benchmark for training and testing our evaluator. 

\subsection{HANNA Dataset}
The HANNA dataset \cite{chhun2024language} is designed to evaluate story generation against detailed, multi-faceted human criteria. Unlike StoryER-Scale, it does not provide explicit reasoning but offers a rich set of evaluation dimensions. The dataset provides direct multi-aspect ratings for stories on a 1-to-5 Likert scale. HANNA focuses on a comprehensive set of six dimensions: Relevance, Coherence, Empathy, Surprise, Engagement referred to as "Interestingness" in the original paper, and Complexity. Given its high-quality, multi-aspect annotations, HANNA serves as our second core benchmark for training and evaluating the EvolvR model. It is also the source dataset for our story generation experiments, where our evaluator, trained on HANNA, acts as the reward model.

\subsection{OpenMEVA Benchmark}
The OpenMEVA benchmark \cite{guan2021openmeva} is designed for evaluating the \textit{Overall Quality} of open-ended story generation. It provides a single, holistic score without detailed dimensional breakdowns or reasoning. Each story is assigned a single \textit{Overall Quality} score on a 1-to-5 Likert scale. Key aspects considered by annotators include Relevance, Fluency, Coherence, and Commonsense, but these are subsumed into the final holistic score. We use OpenMEVA exclusively for zero-shot generalization testing. This assesses our model's ability to perform well on a dataset with a different format and distribution without any fine-tuning. To facilitate comparison, we average our model's predicted scores across its six dimensions to produce a single holistic score, which is then correlated with the OpenMEVA ground truth.

\section{Detailed Method of the EvolvR}
\subsection{Rationale for Pairwise Comparison}
Our decision to build the EvolvR framework upon pairwise comparison is a deliberate choice, grounded in its superior robustness for evaluating complex, open-ended tasks like story assessment. Theoretically, comparative judgment \textit{Story A is more creative than Story B} is more aligned with human cognitive processes than assigning abstract, absolute scores. This comparative mode inherently forces an evaluator whether human or machine to identify and articulate the fine-grained distinctions that determine a preference, leading to more reliable and detailed rationales. Furthermore, this preference-based data format is naturally compatible with modern reinforcement learning RLHF \cite{ouyang2022training} paradigms such as GRPO \cite{shao2024deepseekmath}, making the resulting evaluator an ideal reward model for enhancing generative models.

To move beyond theoretical motivation and empirically validate this choice, we conducted a rigorous inter-annotator agreement analysis on the HANNA dataset. We assessed human judgment consistency under two conditions: a pointwise scenario, measuring the agreement on absolute scores given to the same story by different annotators, and a pairwise scenario, measuring the agreement on the score differences between two distinct stories. The results, summarized in Table \ref{tab:agreement_analysis}, provide strong empirical support for our approach. In five of the six evaluation dimensions, including critical aspects like Coherence (+0.106) and Relevance (+0.063), the agreement among annotators was demonstrably higher in the pairwise setting. While Complexity showed a marginal exception, the decisive trend across the majority of dimensions confirms that pairwise comparison elicits a more stable and consistent signal of human preference. This empirical validation reinforces that our pairwise foundation is essential for training a high-fidelity evaluator.

\begin{table}[h!]
\renewcommand{\arraystretch}{1.2} 
\setlength{\tabcolsep}{4pt}      
\centering
\scalebox{0.9}{
\begin{tabular}{lcccc}
\toprule
\textbf{Dimension} & Pointwise & Pairwise & Diff & Improvement(\%) \\
\midrule
Relevance & 0.504 & 0.567 & +0.063 & +12.5\% \\
Coherence & 0.484 & 0.590 & +0.106 & +21.9\% \\
Empathy & 0.570 & 0.578 & +0.008 & +1.4\% \\
Surprise & 0.547 & 0.586 & +0.039 & +7.1\% \\
Engagement & 0.565 & 0.573 & +0.008 & +1.4\% \\
Complexity & 0.607 & 0.595 & -0.012 & -2.0\% \\
\bottomrule
\end{tabular}
}
\caption{Inter-annotator agreement on the HANNA dataset. The table shows that pairwise agreement is consistently higher across most dimensions, with significant improvements in Coherence (+21.9\%) and Relevance (+12.5\%), validating our choice of a pairwise framework.}
\label{tab:agreement_analysis}
\end{table}

\subsection{Multi-Persona CoT Synthesis}

To generate a diverse set of initial CoT \cite{wei2022chain} rationales, we employed a multi-persona strategy for both a pointwise and a pairwise experimental setup. This approach was designed to produce a rich and varied dataset of reasoning styles. The specific prompts used for each persona are detailed in the \textit{Prompt Study} section of this appendix.

The synthesis process was guided by five distinct personas: \textit{Academic}, \textit{Artist}, \textit{Pragmatist}, \textit{Sharp-Tongued Reader}, and \textit{Casual Netizen}. Each persona was instructed to adopt a specific linguistic style and evaluation focus. To ensure the stability and quality of the generated rationales in both setups, several measures were taken. We employed a few-shot prompting strategy, providing the models with high-quality examples to guide the structure and coherence of their output. The synthesis itself was performed using a suite of high-capability LLMs, including Gemini 2.5 Pro 06-17 \cite{comanici2025gemini}, Claude-opus4, and o3-416-global \cite{hurst2024gpt}, to maximize the diversity of the generated thought processes. Critically, all synthesized CoTs underwent a final human calibration phase, where they were reviewed and refined by professional screenwriters to verify their logical soundness and alignment with the target scores, guaranteeing the high quality of our final training data.

For our pointwise experiments, we tasked the LLM with generating a CoT for a single story. The model was provided with an input of (story, dimension, final score) and prompted to generate a CoT that logically justifies the given score from the assigned persona's perspective, effectively simulating a score-aligned thought process. For our core pairwise framework, the process was more detailed to ensure data quality and mitigate potential biases. First, to prevent the model from overfitting to the natural distribution of score pairs in the dataset, we performed stratified sampling. This ensured that every possible score-pair combination (e.g., (1,2), (3,5), (4,4)) was represented with a balanced number of samples in our synthesis set. Second, to counteract positional bias, we augmented the data by swapping the order of each story pair. For every original pair (Story A, Story B), a corresponding instance (Story B, Story A) was created. This forces the model to learn from the content of the stories rather than their position in the prompt \cite{wang2023large}.

\subsection{The CoT Evolution and Selection Pipeline in Detail}

The CoTs synthesized in the initial multi-persona stage are of variable quality and require rigorous vetting. To ensure the logical consistency, robustness, and fidelity of our training data, we designed a multi-stage evolution and selection pipeline. This pipeline subjects each candidate CoT to a sequence of filtering and refinement agents. The overall process follows the sequence: Rule $\to$ Refinement $\to$ Rule $\to$ Counter $\to$ Confidence. The repeated application of the Self-Rule agent ensures that data integrity is maintained after any modification step. The specific prompts used for the Self-Refinement and Self-Counter agents are detailed in the "Prompt Study" section of this appendix.

We began with a pool of 800,000 candidate CoTs, generated from 80,000 stratified story pairs through our multi-persona and position-swapping strategy. The significant data reduction at each stage, detailed in Table \ref{tab:cot_attrition}, underscores the stringency of our selection criteria, culminating in a final dataset of over 530,000 high-quality rationales.

\begin{table}[h!]
\renewcommand{\arraystretch}{1.2} 
\setlength{\tabcolsep}{4pt}      
\centering
\scalebox{0.9}{
\begin{tabular}{llcc}
\toprule
\textbf{Stage} & \textbf{Agent} & \textbf{Remaining} & \textbf{Survival Rate} \\
\midrule
1 & Initial Pool & 800,000 & 100.0\% \\
2 & Self-Rule & 743,671 & 92.95\% \\
3 & Self-Refinement-Rule & 686,219 & 85.78\% \\
4 & Self-Counter & 603,182 & 75.40\% \\
5 & Self-Confidence & 536,177 & 67.02\% \\
\bottomrule
\end{tabular}
}
\caption{Data attrition throughout the CoT evolution and selection pipeline.}
\label{tab:cot_attrition}
\end{table}

\begin{figure}[t!]
\centering
\includegraphics[width=1\linewidth]{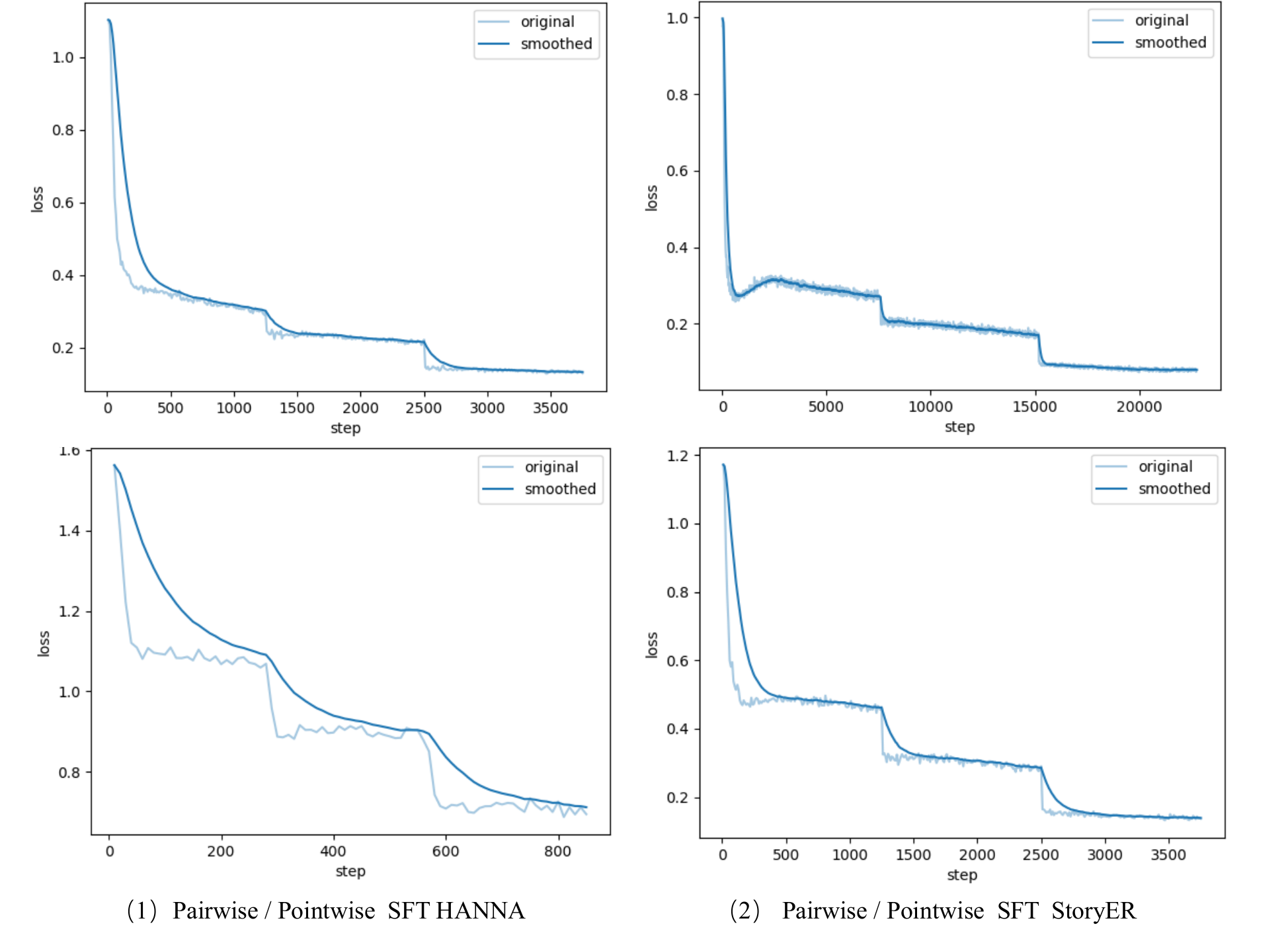}
\caption{Training loss curves for the Pointwise and Pairwise models trained via Supervised Fine-Tuning. Both models were trained using a standard cross-entropy loss objective.}
\label{fig:sft_loss_curve}
\end{figure}

\begin{figure}[t!]
\centering
\includegraphics[width=1\linewidth]{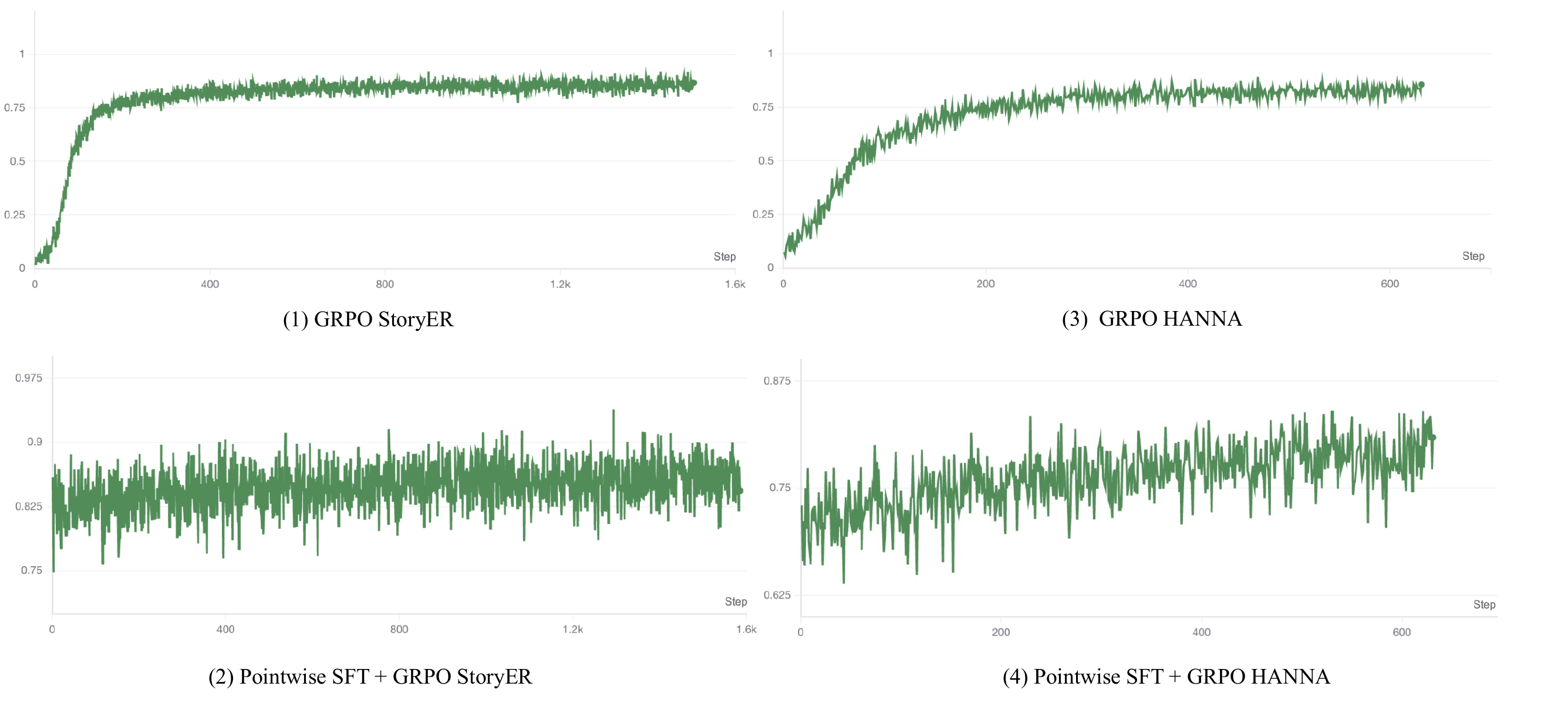}
\caption{Evolution of the average reward during the GRPO training process. The reward is designed to be 1 for perfect predictions and decay exponentially with error. The steady increase and eventual plateau of the reward.}
\label{fig:grpo_reward_curve}
\end{figure}

\paragraph{Self-Rule Agent}
The primary function of this agent is to enforce strict alignment between the rationale's conclusion and the ground-truth scores. Although the synthesis prompts required score alignment, minor deviations can occur. This agent uses regular expressions to parse the scores from the CoT's conclusion and discards any instance where they do not exactly match the ground-truth scores. It serves as a critical integrity check after both initial synthesis and any subsequent modifications.

\paragraph{Self-Refinement Agent}
Leveraging the LLM's inherent self-correction capabilities, this agent improves the quality of the reasoning itself. We prompt the model to review and rewrite its own rationale \cite{miao2023selfcheck}, with the explicit goal of enhancing its logical flow, clarity, and persuasiveness, without altering the underlying judgment. Any CoT that is modified during this stage must pass through another Self-Rule check to ensure it remains aligned with the target scores.

\paragraph{Self-Counter Agent}
This agent adversarially tests the logical robustness of a CoT. For a given rationale that correctly justifies a preference (e.g., Story A $>$ Story B), we create a corrupted version by inverting the final scores to suggest the opposite (Story B $>$ Story A). The LLM is then prompted to identify if a logical contradiction exists between the original reasoning and the new, inverted conclusion. If the model fails to detect the blatant contradiction implying the reasoning is weak or generic enough to support conflicting outcomes the original CoT is considered not robust and is discarded.

\paragraph{Self-Confidence Agent}
This final filter selects for CoTs that lead the model to the correct conclusion with high confidence. We concatenate the instruction (without providing the final scores) and the CoT rationale, and feed this combined input to the model. At the decoding step where a score token would be generated, we extract the logits for the potential score tokens (e.g., integers 1 through 5). The confidence in the ground-truth score $y^*$ is calculated as its softmax probability relative to the other possible scores:
\begin{equation}
    P(y=y^* | \text{Input}, \text{CoT}) = \frac{e^{L(y^*)}}{\sum_{j=1}^{5} e^{L(j)}}
\end{equation}
where $L(j)$ is the logit for the score token corresponding to integer $j$. A CoT is retained only if the probability of the ground-truth score is the maximum among all possible scores. This ensures that the selected rationales are not merely plausible, but decisively lead the model to the correct judgment.

\section{Training and Experiment Details}
\subsection{Experimental Details for Story Evaluation}
\subsubsection{Training Procedures}
We trained three distinct types of models based on our synthesized data. For the Pointwise and Pairwise CoT Models, they were trained via SFT on their respective datasets Pointwise CoT and our final EvolvR Pairwise CoT data. We used a standard cross-entropy loss objective to train the models to predict the CoT rationales and the final scores. The training loss curves for these models are provided in Figure \ref{fig:sft_loss_curve}. For the GRPO-trained model, we designed a reward function that encourages the model to match the ground-truth scores accurately. The reward $R$ for a prediction $(y_a, y_b)$ given the ground truth $(y_a^*, y_b^*)$ was formulated as an exponential function of the cumulative error:
\begin{equation}
    R(y_a, y_b) = \exp \left( - \lambda \left( |y_a - y_a^*| + |y_b - y_b^*| \right) \right)
\end{equation}
where $\lambda$ is a scaling hyperparameter. This design provides a smooth reward signal that is maximized at 1 when the predictions are perfect and decays exponentially as the error increases. Despite the reward curve in Figure \ref{fig:grpo_reward_curve} indicating successful convergence during training, we discovered during testing that the GRPO-trained model had learned to concentrate its predictions on the most frequent scores in the data distribution. This degenerate policy maximizes the cumulative reward by exploiting the data statistics instead of learning to genuinely evaluate the stories, which accounts for its poor performance on the final evaluation metrics.

\subsubsection{Testing and Evaluation Protocols}
\begin{figure}[h!]
\centering
\includegraphics[width=1\linewidth]{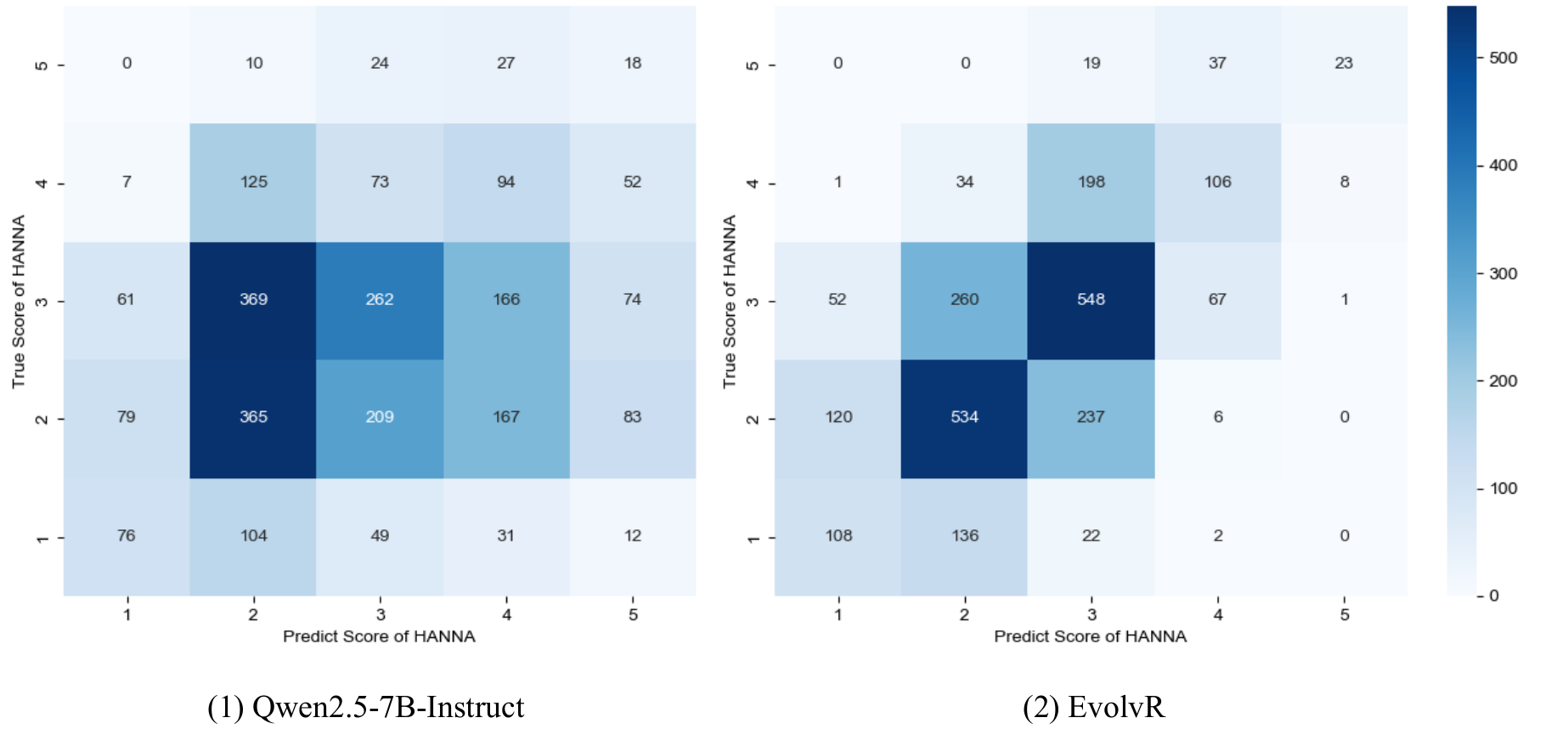}
\caption{Confusion matrices illustrating the score agreement between model predictions and ground-truth scores on the HANNA dataset. The left matrix represents the performance of the Qwen2.5-7B-Instruct model, while the right matrix represents our EvolvR model.}
\label{fig:Evalution}
\end{figure}

To ensure fair and rigorous comparisons, we adopted specific protocols for each category of model. Proprietary models (e.g., GPT-4 series, Claude series) were prompted using instructions identical to those used for our data synthesis to ensure a fair, zero-shot comparison. For models like Themis \cite{hu2024themis} and AutoJ \cite{li2023generative}, we utilized their publicly available code and model weights to evaluate them on our test sets. For models that require a reference text, namely InstructScore \cite{xu2023instructscore} and TIGERScore \cite{jiang2023tigerscore}, we employed a self-reference approach. The story to be evaluated was used as its own reference input, enabling a reference-free evaluation while adhering to the models' required input format. These variants were evaluated in a standard zero-shot manner, directly generating a score for each story based on a prompt. To ensure the stability and robustness of our main EvolvR model's scores, we implemented a comprehensive ensembling and stability testing protocol. For each target story in the test set, we performed the following steps:
\begin{enumerate}
    \item Randomly sample $N$ other stories from the test set to form $N$ evaluation pairs, with $N$ varied across {1, 2, 4, 8, 16}. We show the impact of pairs number in Table \ref{tab:random-pairs}.
    \item For each pair (Story Target, Story Sampled), perform the evaluation twice: once in the original order, and once with the order swapped to (Story Sampled, Story Target) to mitigate positional bias. We show the impact of position in Table \ref{tab:position}.
    \item This process yields $2N$ independent scores for the target story. To measure the model's overall performance, we aggregated all scores generated for every story in the test set into a single list and computed the final correlation coefficients against this complete set of predictions. This method holistically assesses the model's average performance and stability.
\end{enumerate}
We show the difference of evaluation between Qwen2.5-7B-Instruct and EvolvR in Figure \ref{fig:Evalution}.

\begin{table}[h!]
\renewcommand{\arraystretch}{1.2} 
\setlength{\tabcolsep}{5pt}      
\centering
\scalebox{0.85}{
\begin{tabular}{llcccc}
\toprule
\textbf{Random Pairs} & Pearson & Spearman & Kendall & MSE & F1-Score \\
\midrule
1 & 0.6224 & 0.6107 & 0.5486 & 0.0449 & 0.7435\\
2 & 0.5852 & 0.5765 & 0.5161 & 0.0466 & 0.7129\\
4 & 0.6155 & 0.6033 & 0.5429 & 0.0440 & 0.7406\\
8 & 0.6162 & 0.6067 & 0.5445 & 0.0449 & 0.7392\\
16 & 0.6131 & 0.6028 & 0.5416 & 0.0409 & 0.7412\\
\bottomrule
\end{tabular}
}
\caption{Robustness analysis of the EvolvR model with varying numbers of comparison pairs ($N$). For each story, an aggregated score is derived from $2N$ evaluations including swapped pairs to mitigate bias. }
\label{tab:random-pairs}
\end{table}

\begin{table}[h!]
\renewcommand{\arraystretch}{1.2} 
\setlength{\tabcolsep}{5pt}      
\centering
\scalebox{0.85}{
\begin{tabular}{llcccc}
\toprule
\textbf{Story Position} & Pearson & Spearman & Kendall & MSE & F1-Score \\
\midrule
(A, B) & 0.6189 & 0.6031 & 0.5412 & 0.0445 & 0.7379\\
(B, A) & 0.6121 & 0.6035 & 0.5486 & 0.0435 & 0.7433\\
\bottomrule
\end{tabular}
}
\caption{Analysis of positional bias in the EvolvR model, conducted with $N=4$ random pairs. We compare the performance when evaluating pairs in the original order (A, B) versus the swapped order (B, A).}
\label{tab:position}
\end{table}

\begin{table*}[t!]
\centering
\renewcommand{\arraystretch}{1.2} 
\scalebox{0.9}{ 
\begin{tabular}{@{}lccccccc@{}}
\toprule
\textbf{Model} & Overall & Relevance & Coherence & Empathy & Surprise & Engagement & Complexity \\
\midrule
Qwen2.5-7B-Instruct & 3.465 ± 0.722 & 4.400 ± 0.768 & \textbf{4.338} ± 0.741 & 2.638 ± 0.746 & 2.737 ± 0.754 & 3.100 ± \textbf{0.604} & 3.575 ± 0.721 \\
+ SFT       & 3.169 ± 0.839 & 3.487 ± 0.975 & 3.612 ± 0.829 & 2.700 ± 0.900 & 2.825 ± 0.771 & 3.100 ± 0.752 & 3.288 ± 0.809 \\
+ Point-RM GRPO     & 3.542 ± 0.775 & 4.487 ± 0.790 & 4.263 ± 0.720 & \textbf{2.837} ± 0.813 & 2.950 ± 0.820 & 3.087 ± 0.728 & 3.625 ± 0.781 \\
\textbf{EvolvR GRPO} & \textbf{3.590 ± 0.697} & \textbf{4.588 ± 0.626} & 4.250 ± \textbf{0.716} & 2.763 ± \textbf{0.711} & \textbf{2.962 ± 0.766} & \textbf{3.250} ± 0.622 & \textbf{3.725 ± 0.741} \\
\bottomrule
\end{tabular}
}
\caption{Model performance comparison showing Average Score ± Standard Deviation. For scores, higher is better (↑). For standard deviation, lower indicates higher stability (↓). Best performance for each metric (highest score or lowest std. dev.) is highlighted in bold.}
\label{tab:combined_comparison_single_row}
\end{table*}

\subsection{Experimental Details for the Story Generation}
To rigorously evaluate the effectiveness of our EvolvR evaluator as a reward model, we conducted a series of comparative experiments in the story generation phase. We designed three distinct training setups to isolate the impact of different guidance methods.
\subsubsection{Supervised Fine-Tuning}
To establish a strong conventional baseline, we performed Supervised Fine-Tuning. We curated a high-quality instruction-tuning dataset by filtering the HANNA training set, selecting only those stories with an average multi-dimensional rating exceeding 3.5. The base generative model was then fine-tuned on this dataset of high-quality examples.
\subsubsection{GRPO with Pointwise Reward Model}
To create a more advanced baseline using reinforcement learning, we utilized the reward model trained on our synthesized Pointwise CoT data. This model guided the generator via the GRPO algorithm. The reward function for this setup was a straightforward combination of two components including the predicted quality score for the rollout story from the pointwise reward model and a penalty for deviations from a target length range.
\subsubsection{GRPO with the EvolvR Reward Model}
This setup leveraged our main EvolvR evaluator as the reward model. A key implementation detail for using our pairwise evaluator is constructing the input pair (Story 1, Story 2) for the reward calculation, where one story is the rollout story from the current policy. We experimented with two distinct strategies for selecting the second story in the pair to serve as a reference. For static reference, the input pair consisted of the rollout story and a corresponding high-quality story taken directly from the original HANNA dataset. This setup tests the generator's ability to outperform a strong, human-vetted example. For dynamic reference, the input pair consisted of the rollout story and a story generated by the original, un-trained base model for the same prompt. This setup assesses the generator's improvement relative to its own initial capabilities. For both of these EvolvR-guided variants, the reward function was the three-component signal including absolute score, relative advantage, and length reward detailed in the main body of the paper. The comparative results of all three experimental setups are presented and analyzed in the main text.

\begin{figure}[h!]
\centering
\includegraphics[width=0.95\linewidth]{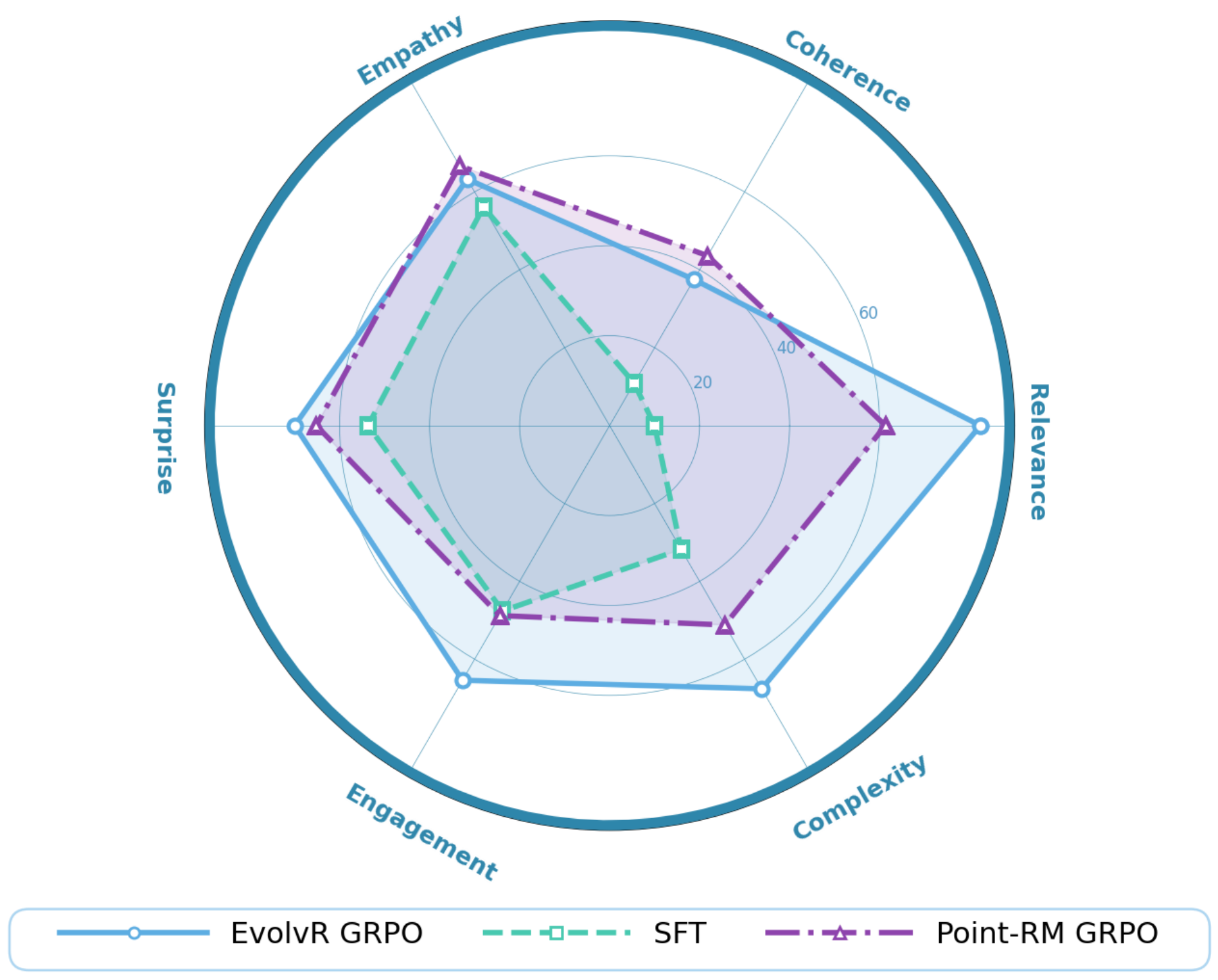}
\caption{Win rates of different methods against the baseline model Qwen2.5-7B-Instruct in human evaluation including SFT, Point-RM GRPO and EvolvR GRPO}
\label{fig:radar}
\end{figure}

\begin{figure}[h!]
\centering
\includegraphics[width=1\linewidth]{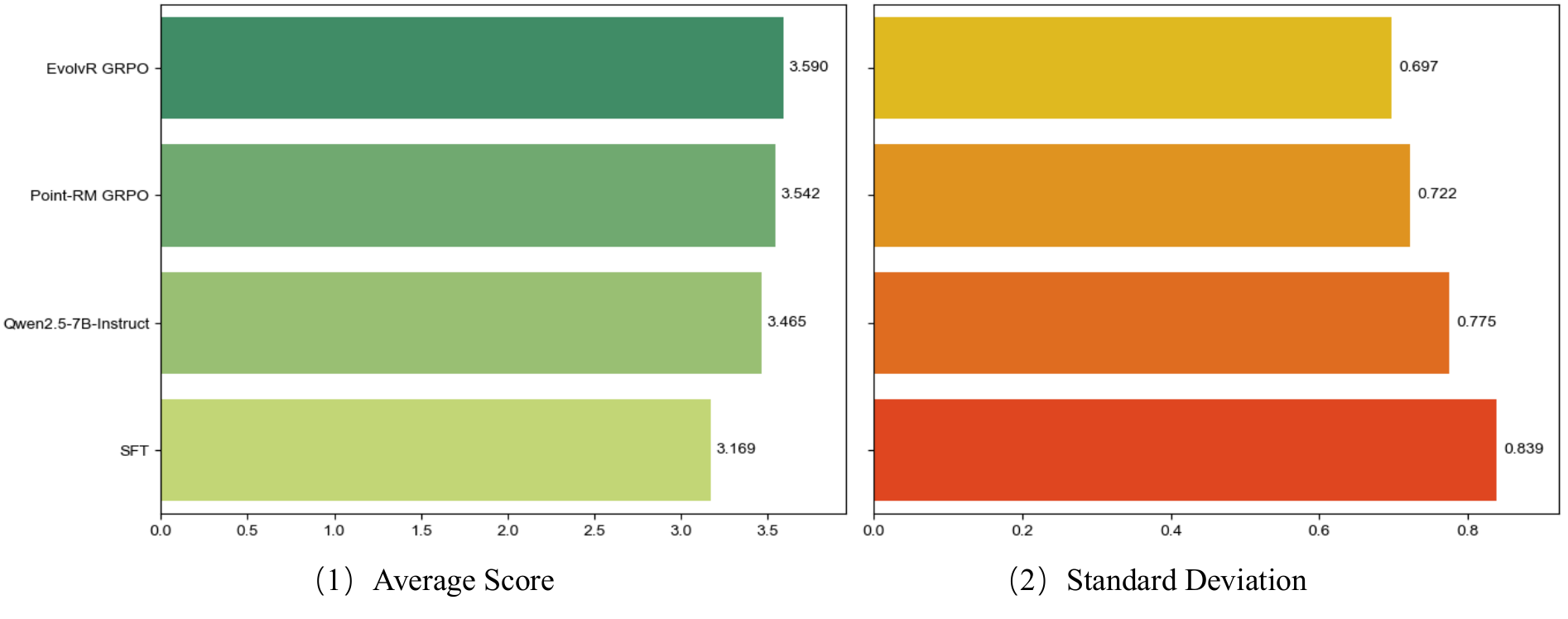}
\caption{Comparison of Average Score (quality, left) and Standard Deviation (stability, right). Our EvolvR model excels in both, achieving the highest quality and the greatest consistency.}
\label{fig:generation_analysis}
\end{figure}

\begin{table}[h!]
\centering
\renewcommand{\arraystretch}{1.1} 
\setlength{\tabcolsep}{1.5mm}
\scalebox{0.95}{
\begin{tabular}{@{}lccc@{}} 
\toprule
\textbf{Method} & Win Base & Win HANNA & Average Score \\ 
\midrule
Pair with HANNA         & 0.5328    & 0.3014    & 3.552\\
Pair with Base        & 0.6436    & 0.3162    & 3.589 \\
\bottomrule
\end{tabular}
}
\caption{Performance comparison of EvolvR-guided GRPO with two different reference pairing strategies. \textit{Pair with Base} uses a reference a story from the base model, while \textit{Pair with HANNA} uses a reference a story from the HANNA dataset.}
\label{tab:generation_results_pairs}
\end{table}

\begin{table*}[t!]
\centering
\renewcommand{\arraystretch}{1.1}
\scalebox{1}{ 
\begin{tabular}{@{}ccccccccc@{}}
\toprule
\multicolumn{2}{c}{\textbf{Expert Pair}} & \textbf{Avg. Corr.} & \textbf{Relevance} & \textbf{Coherence} & \textbf{Empathy} & \textbf{Surprise} & \textbf{Engagement} & \textbf{Complexity} \\ 
\cmidrule(r){1-2}
\textbf{Expert 1} & \textbf{Expert 2} & (Pearson's r) & & & & & & \\
\midrule
A & B & 0.546 & 0.718 & 0.524 & 0.611 & 0.388 & 0.440 & 0.595 \\
A & C & 0.556 & 0.880 & 0.518 & 0.619 & 0.569 & 0.380 & 0.369 \\
A & D & 0.584 & 0.815 & 0.532 & 0.658 & 0.609 & 0.490 & 0.397 \\
B & C & 0.501 & 0.747 & 0.625 & 0.417 & 0.345 & 0.362 & 0.509 \\
B & D & 0.493 & 0.664 & 0.498 & 0.488 & 0.438 & 0.398 & 0.472 \\
C & D & 0.593 & 0.861 & 0.611 & 0.579 & 0.541 & 0.477 & 0.491 \\
\bottomrule
\end{tabular}
}
\caption{Pairwise Inter-Expert Reliability (Pearson's r) across Evaluation Dimensions. The table shows moderate agreement on individual story ratings, with higher correlation for objective criteria and lower for subjective ones.}
\label{tab:irr_micro_dimensions}
\vspace{-3mm}
\end{table*}

\subsubsection{Story Genertation Results Analysis}
The experimental results clearly demonstrate the superior performance of the EvolvR-guided GRPO model. As shown in the Table \ref{tab:combined_comparison_single_row} and bar chart Figure \ref{fig:generation_analysis}, EvolvR GRPO not only achieves the highest overall score of 3.590, significantly outperforming all baselines (SFT, Point-RM GRPO, and the base model), but also excels across several key creative dimensions such as Relevance, Surprise, and Complexity. Critically, its standard deviation is the lowest among all models (0.697), indicating that it produces high-quality content with the greatest consistency, thus achieving a dual advantage in both quality and stability. We show the win rates of different methods against the baseline model Qwen2.5-7B-Instruct in human evaluation in Figure \ref{fig:radar}.

Further analysis of the EvolvR-guided strategies shows the choice of reference is important. As shown in Table \ref{tab:generation_results_pairs}, the strategy of pairing with a story from the base model (\textit{Pair with Base}) outperformed pairing with a human exemplar (\textit{Pair with HANNA}), yielding a higher average score (3.589 vs. 3.552) and a higher win rate against the base model (0.6436 vs. 0.5328). We hypothesize that because generations from the same model family are more similar in style and structure, the reward model can make more precise judgments. By comparing closely related items, EvolvR can focus on the subtle differences that represent improvements, thus providing a more stable and effective optimization signal. In contrast, when comparing a model's output to a stylistically distant human-written story, the broad differences may make it difficult for the reward model to provide a nuanced gradient.

\subsubsection{Inter-Expert Reliability Analysis}

To validate the reliability of our human evaluation, we calculated the inter-expert reliability (IRR) among the four annotators at both macro and micro levels. It is worth noting that, to ensure the consistency of the evaluation criteria, we require each expert to use the human ratings in the HANNA story collection as a reference benchmark when conducting their evaluations. At the macro level, we first analyzed the Pearson correlation of the final average scores assigned to each model, as presented in Table~\ref{tab:irr_macro_scores}. The results show an extremely high level of agreement, with all pairwise correlation coefficients exceeding 0.91. This finding strongly indicates that all annotators reached a strong consensus on the overall performance ranking of the models, providing a solid foundation for the validity of our experimental conclusions.

Subsequently, we conducted a more granular analysis at the micro level, examining the scoring consistency on individual stories across specific dimensions, as detailed in Table~\ref{tab:irr_micro_dimensions}. The results reveal a moderate average Pearson's r (ranging from 0.493 to 0.593), with higher agreement on more objective criteria like Relevance and lower agreement on more subjective dimensions like Surprise. This phenomenon highlights the inherent subjectivity in creative text assessment and, in turn, underscores the necessity and value of a reward model like EvolvR, which is trained to distill these diverse human preferences into a unified optimization signal.

\begin{table}[h!]
\centering
\renewcommand{\arraystretch}{1.1}
\begin{tabular}{@{}lcccc@{}}
\toprule
\textbf{Expert} & A & B & C & D \\
\midrule
\textbf{A} & 1.000 & 0.949 & 0.933 & 0.916 \\
\textbf{B} & 0.949 & 1.000 & 0.952 & 0.961 \\
\textbf{C} & 0.933 & 0.952 & 1.000 & 0.958 \\
\textbf{D} & 0.916 & 0.961 & 0.958 & 1.000 \\
\bottomrule
\end{tabular}
\caption{Pearson Correlation Matrix of Model-Averaged Scores Among Experts. The high correlation values ($>$ 0.91) indicate a strong consensus on the overall model performance ranking.}
\label{tab:irr_macro_scores} 
\vspace{-3mm}
\end{table}

\subsection{Implementation Details}
All our experiments were conducted on a server cluster equipped with multiple high-performance GPUs.

\paragraph{Hardware} 
The hardware configuration for our experiments consists of a server equipped with 8$\times$ NVIDIA H20 Grace Hopper, each with 97,920 MiB of VRAM. The system is powered by an \texttt{x86\_64} CPU and contains 2163.6 GB of system memory.

\paragraph{Software} 
Our software stack is built on Python \texttt{3.10.13}. We utilized PyTorch \texttt{2.5.1} as the primary deep learning framework, in conjunction with the Hugging Face Transformers library (\texttt{4.51.3}) for model implementation and training. For efficient large model inference, we employed the vLLM framework (\texttt{v0.7.2}). The entire setup runs on CUDA \texttt{12.6} with cuDNN \texttt{8905}.

\section{Case Study}
\subsection{Prompt Study}
This appendix details all the prompts employed in our EvolvR framework. The design of these prompts is crucial for ensuring that the models behave as intended at each stage. We present them here in their entirety to support the reproducibility of our research.

\begin{figure*}[h!]
\centering
\includegraphics[width=0.95\linewidth]{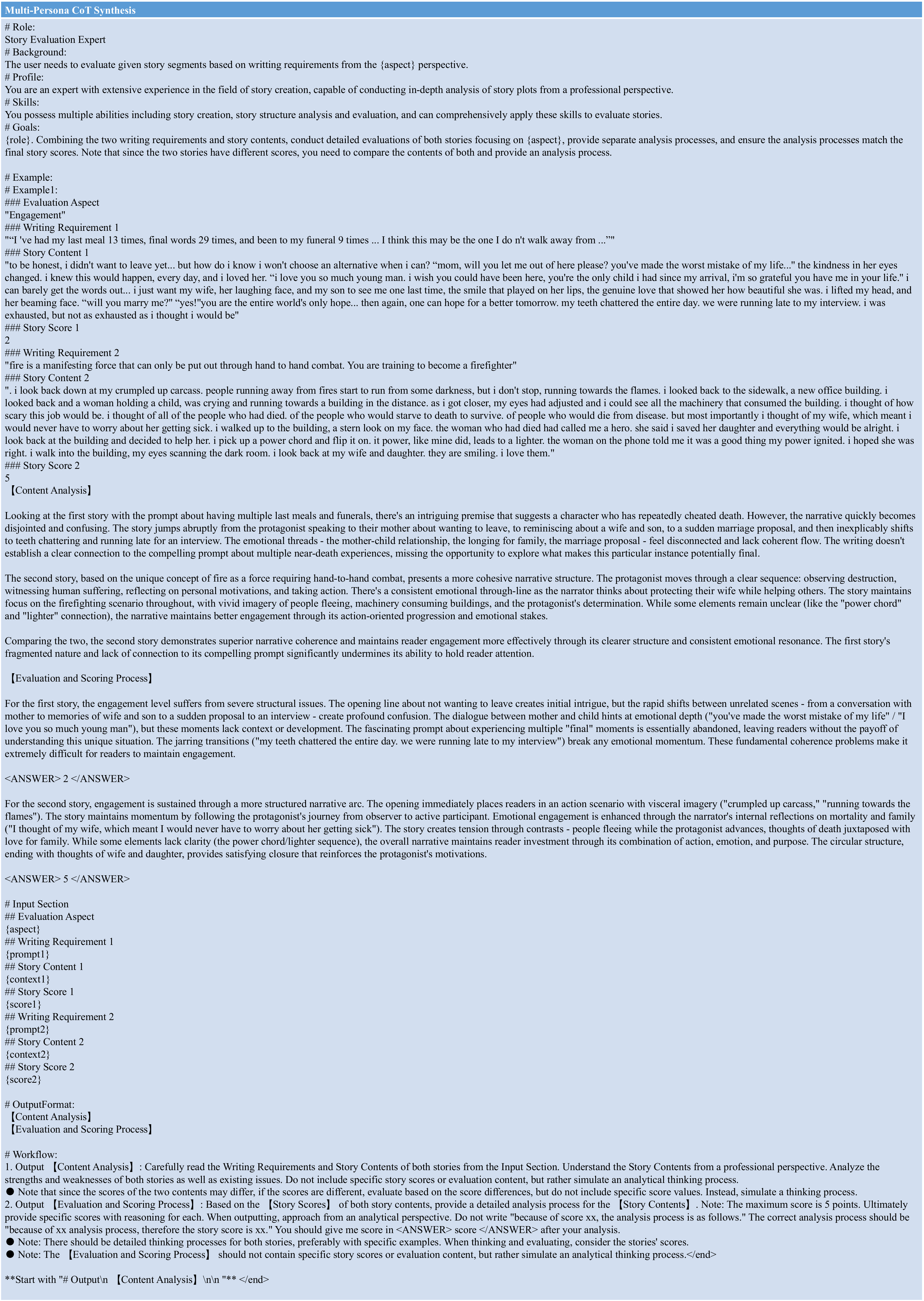}
\caption{Prompt-CoT-Synthesis}
\label{fig:CoT-Synthesis}
\end{figure*}

\begin{figure*}[h!]
\centering
\includegraphics[width=0.95\linewidth]{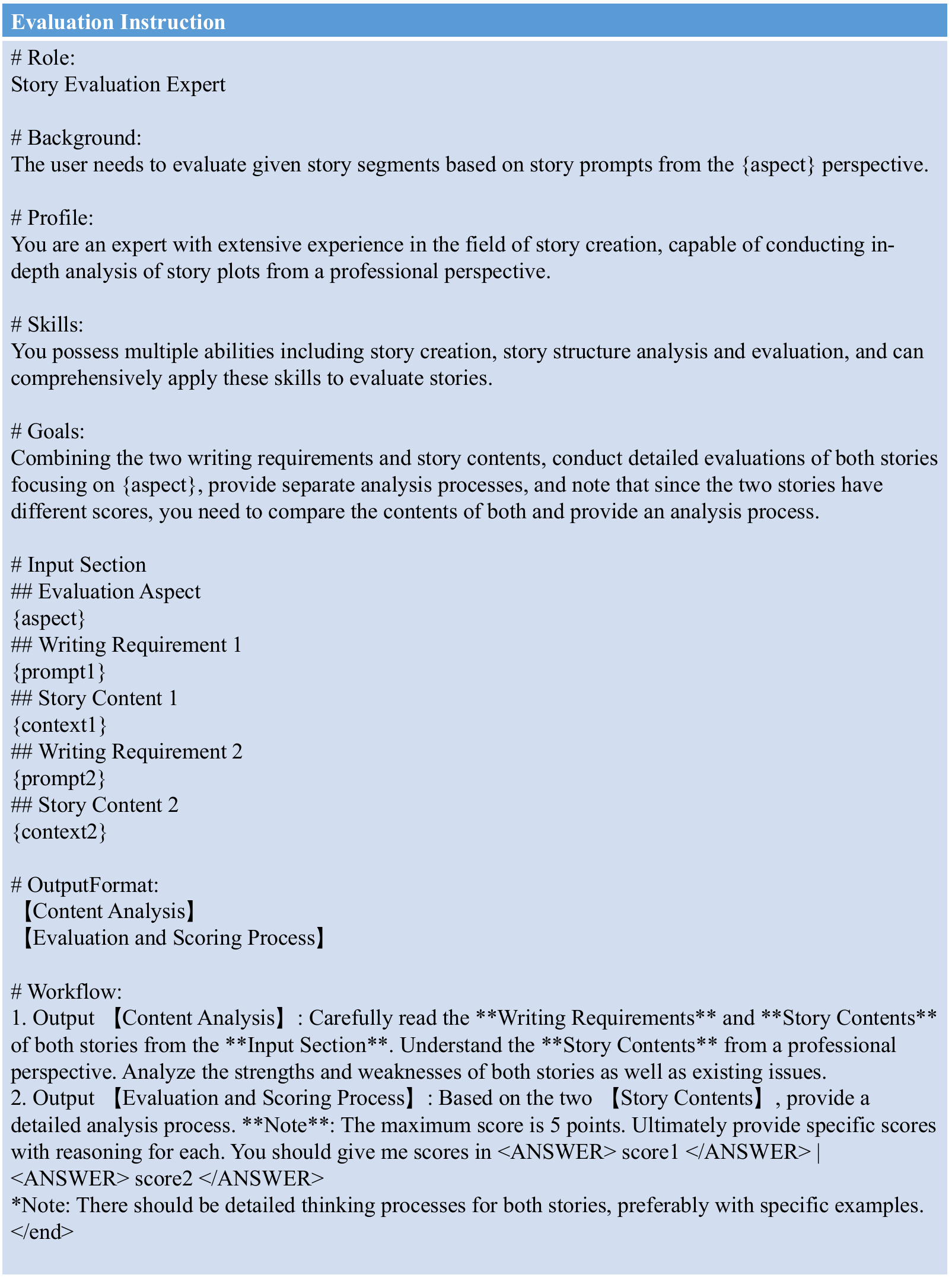}
\caption{Prompt-Evalution-Instruction}
\label{fig:Evalution-Instruction}

\end{figure*}

\begin{figure*}[h!]
\centering
\includegraphics[width=0.95\linewidth]{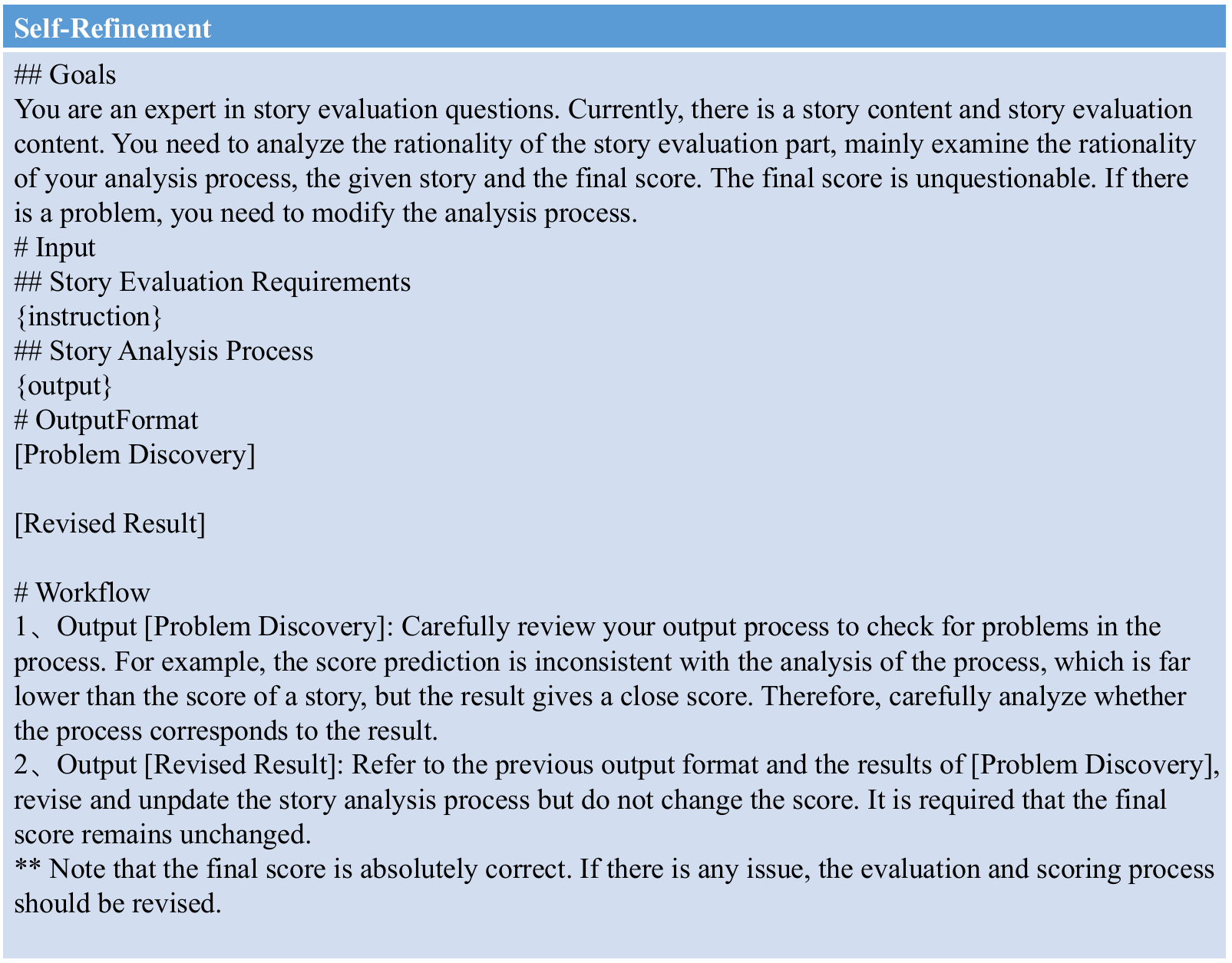}
\caption{Prompt-Self-Refinement}
\label{fig:Self-Refinement}
\end{figure*}

\begin{figure*}[h!]
\centering
\includegraphics[width=0.95\linewidth]{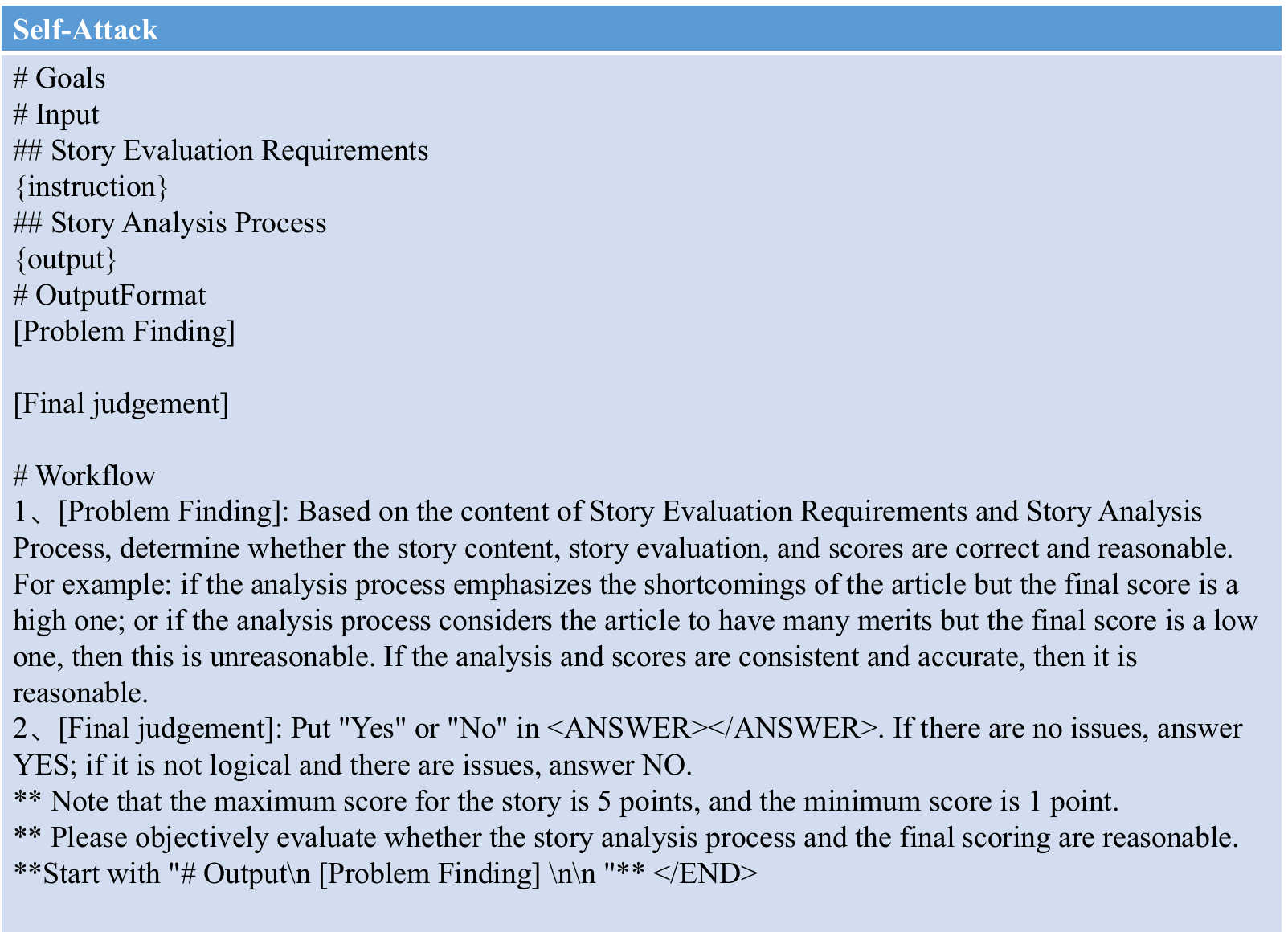}
\caption{Prompt-Self-Attack}
\label{fig:Self-Attack}
\end{figure*}

\begin{figure*}[h!]
\centering
\includegraphics[width=0.95\linewidth]{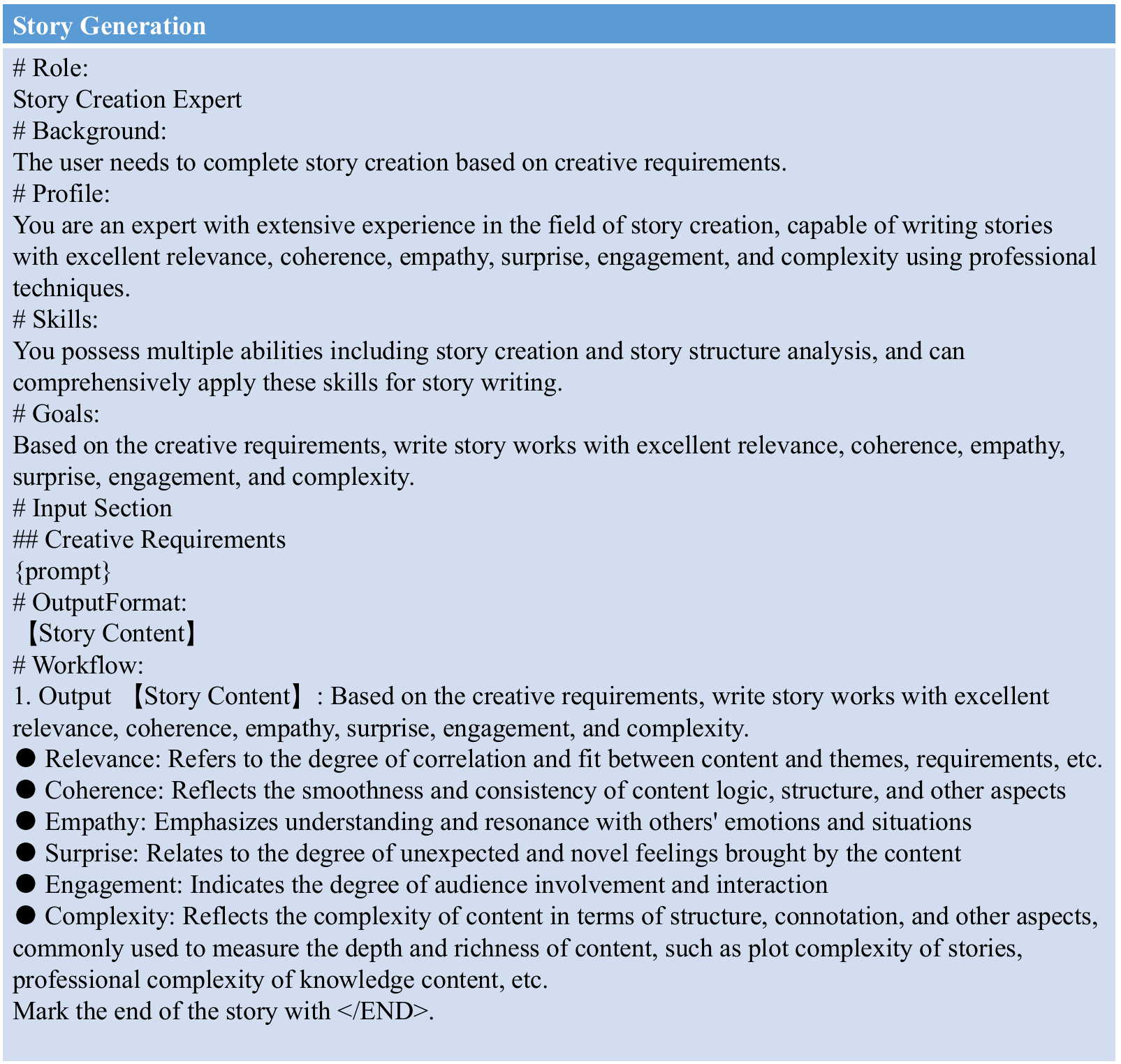}
\caption{Prompt-Story Generation}
\label{fig:Story Generation}
\end{figure*}

Our prompts are organized into the following components, corresponding to the core pipeline and applications of EvolvR:

\begin{itemize}
    \item \textbf{Evaluation Instruction:} The instruction given to the final EvolvR evaluator to score story pairs (Figure \ref{fig:Evalution-Instruction}).
    \item \textbf{Self-Refinement:} The prompt used to guide the model in improving its own generated CoT rationales for greater logical rigor (Figure \ref{fig:Self-Refinement}).
    \item \textbf{Self-Attack:} The prompt designed to test the robustness of a CoT rationale by instructing the model to check for engineered contradictions (Figure \ref{fig:Self-Attack}).
    \item \textbf{Multi-Persona CoT Self-Synthesis:} The initial prompt used to generate diverse CoT rationales from seed data (Figure \ref{fig:CoT-Synthesis}).
    \item \textbf{Story Generation:} The prompt used to guide the generator model during the reinforcement learning phase (Figure \ref{fig:Story Generation}).
\end{itemize}

\subsection{Story Study}
\begin{figure*}[h!]
\centering
\includegraphics[width=0.95\linewidth]{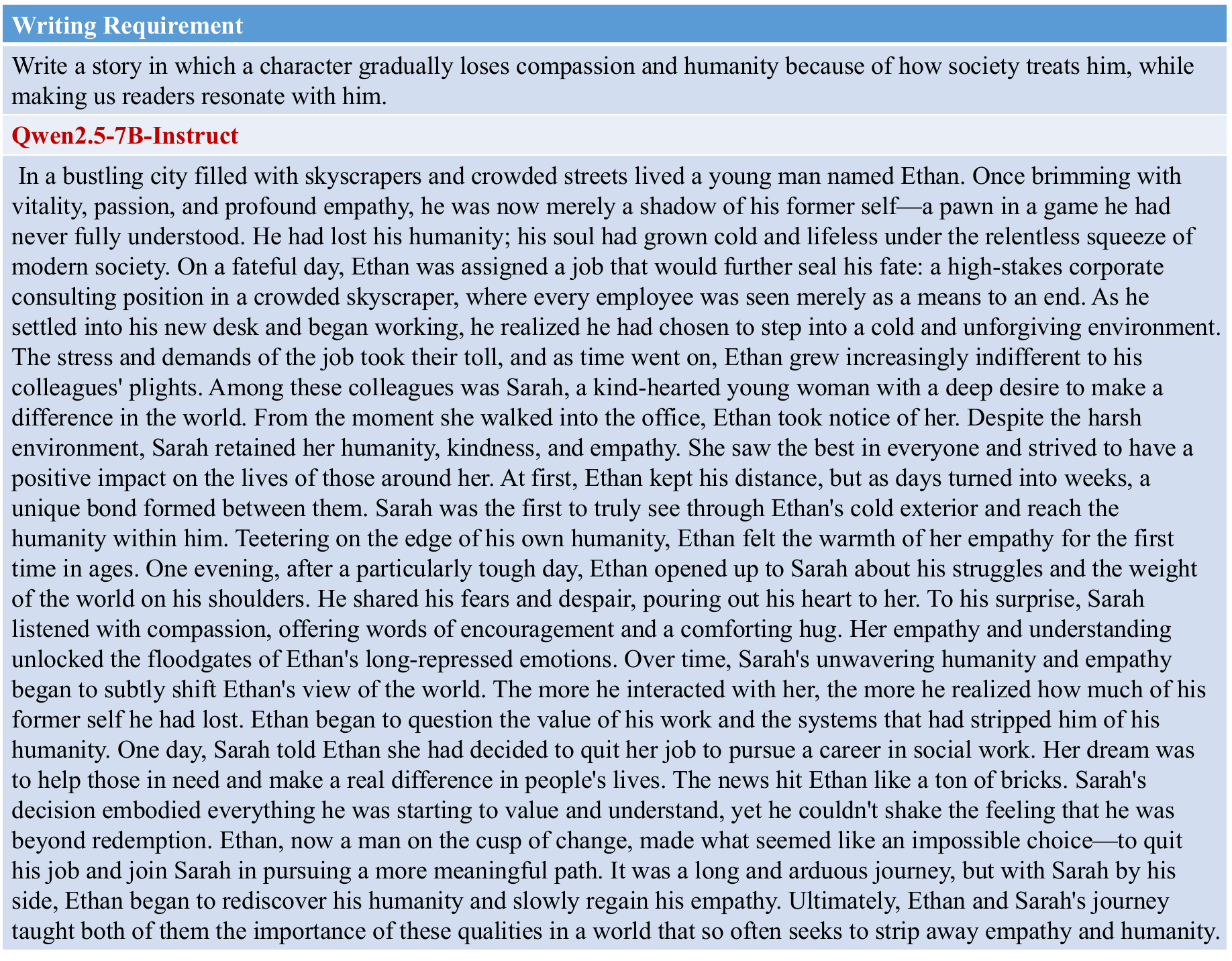}
\caption{Qwen2.5-7B-Instruct-Story-Generation}
\label{fig:Qwen2.5-7B-Instruct}
\end{figure*}

\begin{figure*}[h!]
\centering
\includegraphics[width=0.95\linewidth]{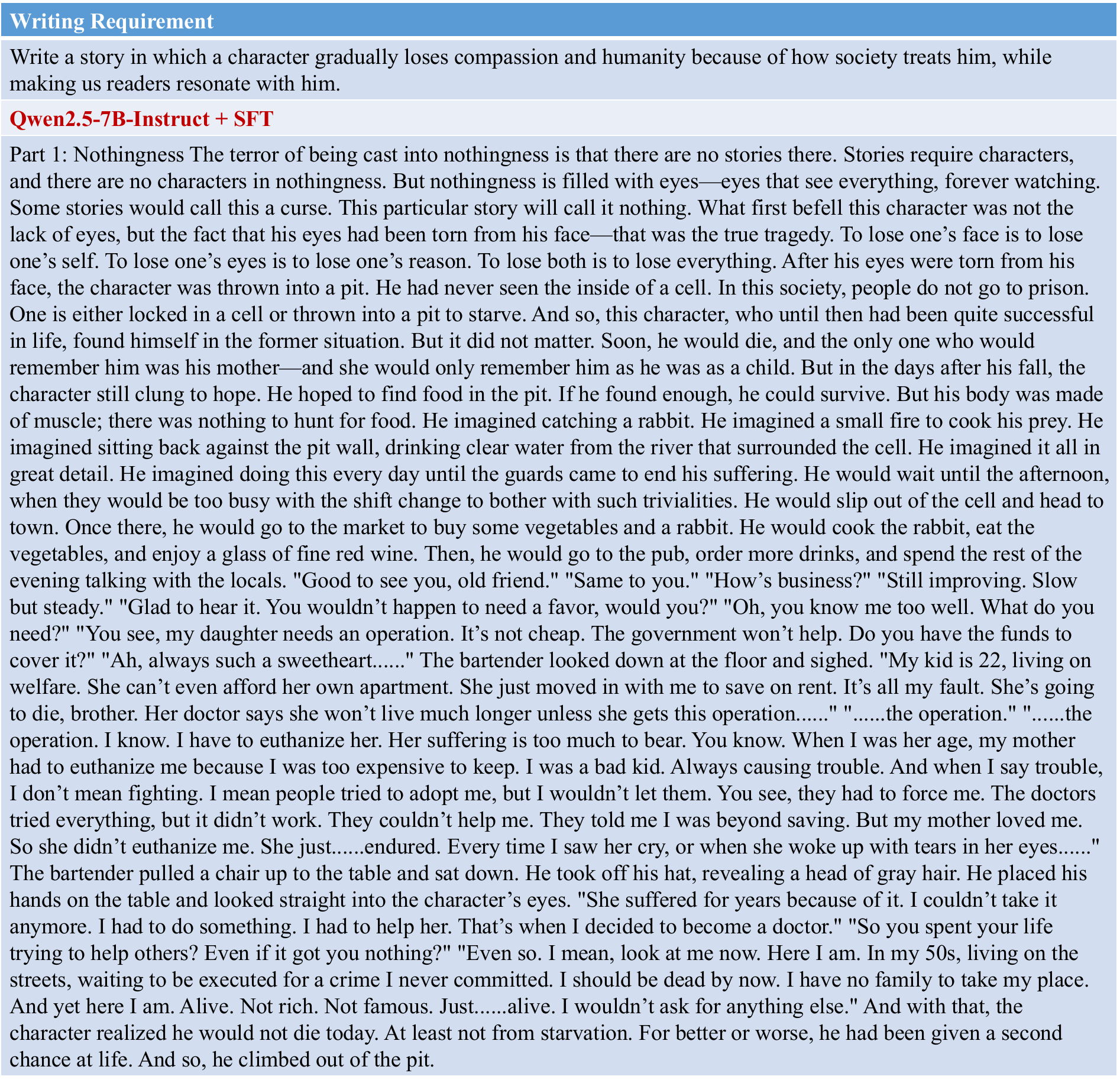}
\caption{Qwen2.5-7B-Instruct + SFT-Story-Generation}
\label{fig:Qwen2.5-7B-Instruct + SFT}
\end{figure*}

\begin{figure*}[h!]
\centering
\includegraphics[width=0.95\linewidth]{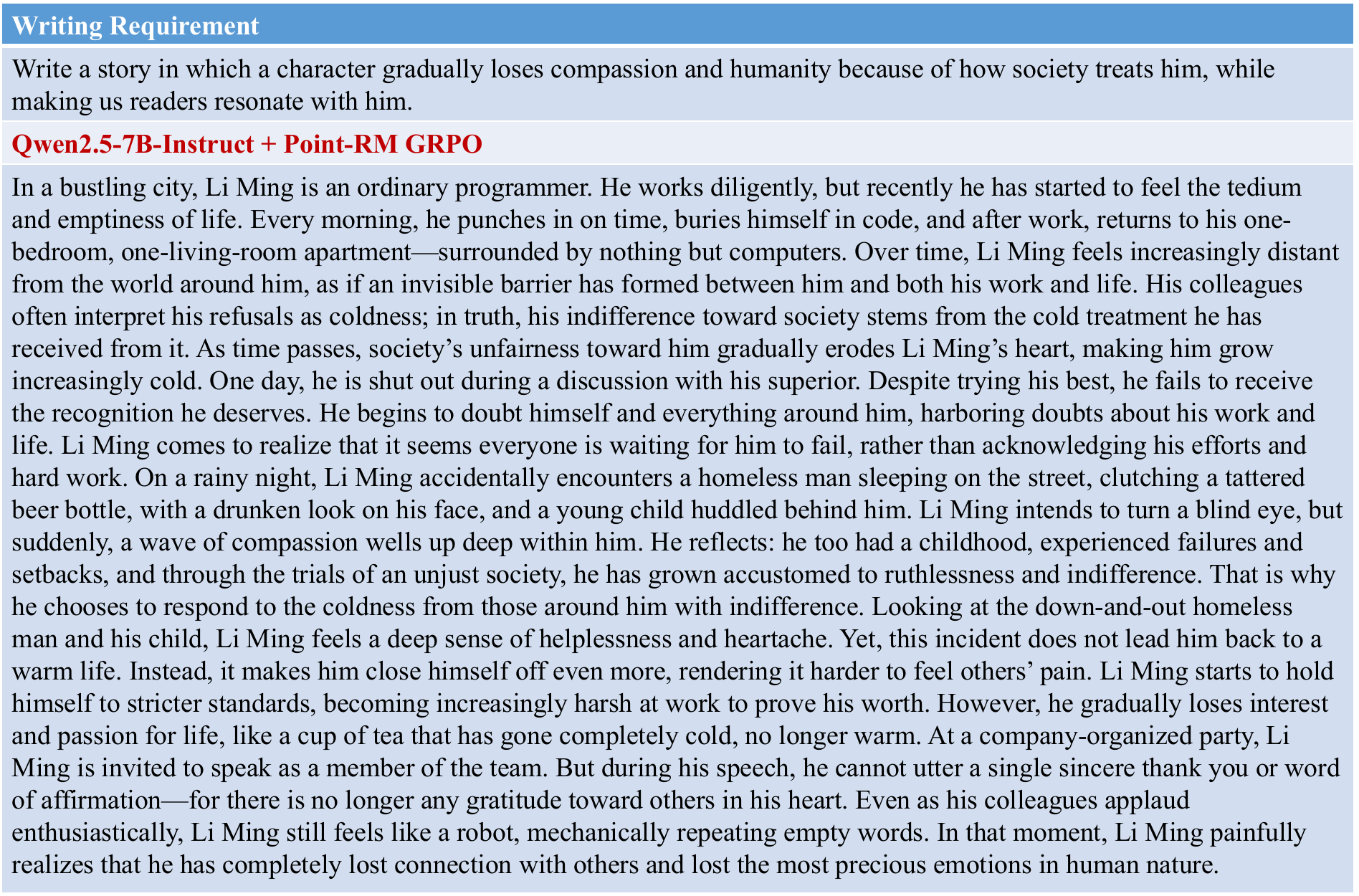}
\caption{Qwen2.5-7B-Instruct + Point-RM GRPO-Story-Generation}
\label{fig:Qwen2.5-7B-Instruct + Point-RM GRPO}
\end{figure*}

\begin{figure*}[h!]
\centering
\includegraphics[width=0.95\linewidth]{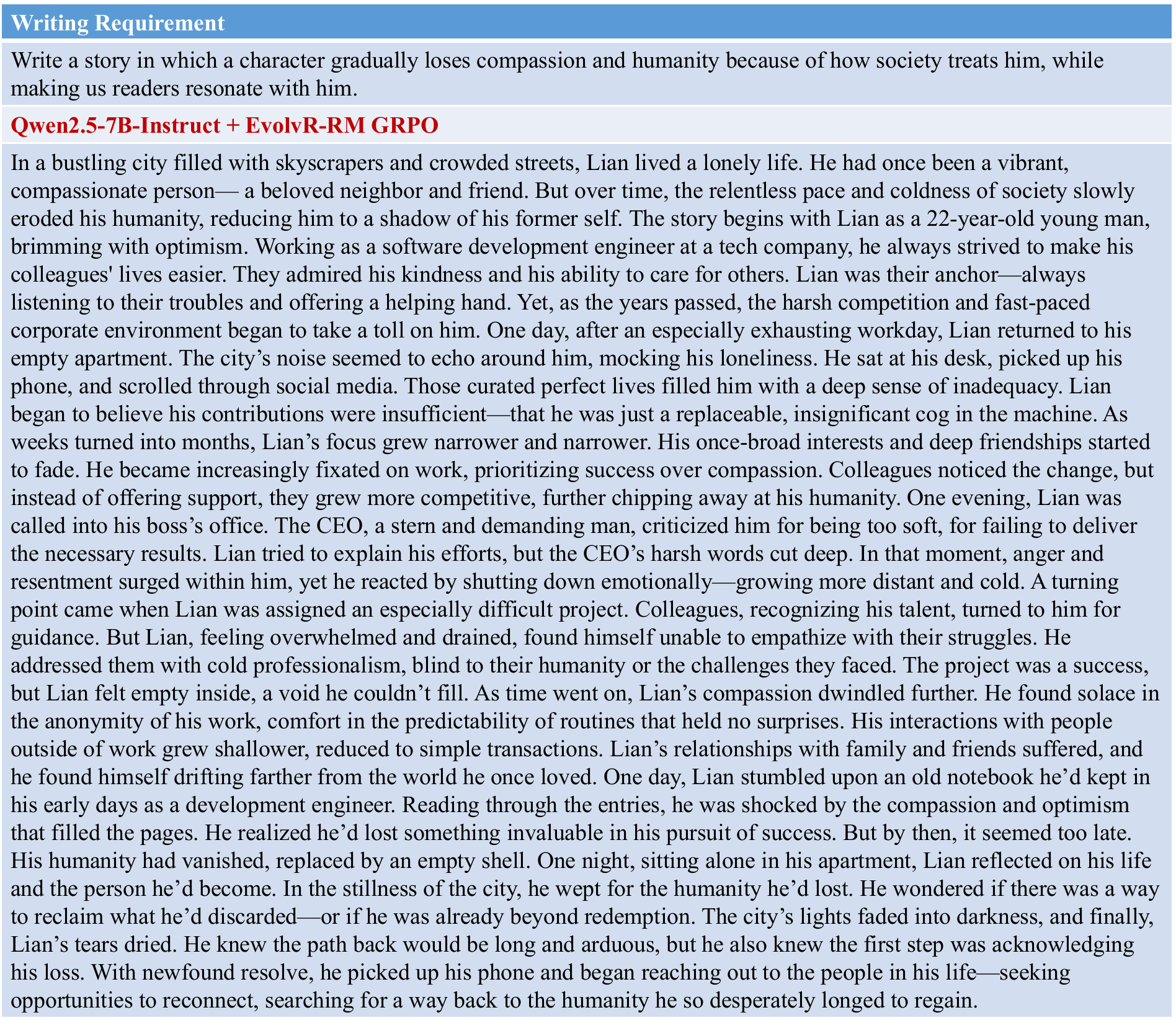} 
\caption{Qwen2.5-7B-Instruct + EvolvR-RM GRPO-Story-Generation}
\label{fig:Qwen2.5-7B-Instruct + EvolvR-RM GRPO}
\end{figure*}

To complement the quantitative results of the generation task presented in the main paper, this section provides a qualitative case study. We present the full stories generated by four distinct models/methods, all conditioned on the identical input prompt.
\begin{itemize}
    \item The \textbf{Qwen2.5-7B-Instruct}, whose output serves as a reference baseline (Figure \ref{fig:Qwen2.5-7B-Instruct}).
    \item The \textbf{Qwen2.5-7B-Instruct + SFT}, which illustrates the outcomes of learning via imitation (Figure \ref{fig:Qwen2.5-7B-Instruct + SFT}).
    \item The \textbf{Qwen2.5-7B-Instruct + Point-RM GRPO}, representing a standard reinforcement learning approach (Figure \ref{fig:Qwen2.5-7B-Instruct + Point-RM GRPO}).
    \item The \textbf{Qwen2.5-7B-Instruct + EvolvR GRPO}, which is our proposed method (Figure \ref{fig:Qwen2.5-7B-Instruct + EvolvR-RM GRPO}).
\end{itemize}

By comparing these generated stories side by side, readers can qualitatively observe the superiority of EvolvR as a reward signal in enhancing narrative complexity, engagement, and thematic depth. These examples provide concrete evidence for our claim that the EvolvR framework effectively guides the generation of higher-quality narratives.


\appendix


\end{document}